\documentclass[runningheads]{llncs}

\usepackage[utf8]{inputenc}
\usepackage[english]{babel}
\usepackage{csquotes}
\usepackage{anyfontsize}
\usepackage{amsmath,mathtools,amsbsy,amssymb,dsfont}
\usepackage{xcolor}
\usepackage{graphicx}
\usepackage[inline]{enumitem}
\usepackage{relsize}
\newlist{compactdescription}{description}{1}\setlist[compactdescription]{noitemsep, leftmargin=2ex}
\usepackage{siunitx}
\usepackage{placeins}

\usepackage[accsupp]{axessibility}  

\usepackage[pagebackref,breaklinks,colorlinks]{hyperref}

\usepackage{tabulary,tabularx,booktabs,multirow}
\robustify\bfseries\newcommand*{\bfs}{\bfseries}

\newcommand*{\themainpaper}{the main paper}

\newcommand*{\forexample}{e.g.\ }

\newcommand*{\idest}{i.e.\ }

\newcommand*{\etal}{et al.\ }

\newcommand*{\binary}[1]{\underline{#1}}
\newcommand*{\booleanop}[1]{\boldsymbol{#1}}
\newcommand*{\Pred}[1]{\text{\smaller\texttt{#1}}}
\newcommand*{\bPred}[1]{\binary{\Pred{#1}}}
\newcommand*{\isPred}[1]{\Pred{Is}_{\text{\Concept{#1}}}}
\newcommand*{\constant}[1]{\Pred{#1}}
\newcommand*{\Concept}[1]{{\smaller\textsf{#1}}}
\newcommand*{\thresh}[1]{\constant{t}_{\text{#1}}}
\newcommand*{\attunedthresh}[1]{\ensuremath{\text{#1}_{\text{b}}}}
\newcommand*{\atoptthresh}[1]{\ensuremath{\overline{\text{#1}}}}
\newcommand*{\tth}[1]{\attunedthresh{#1}}
\newcommand*{\ol}[1]{\atoptthresh{#1}}

\DeclareMathOperator*{\mean}{mean}

\usepackage{subcaption}
\usepackage[capitalize]{cleveref}
\crefname{section}{Sec.}{Secs.}
\Crefname{section}{Section}{Sections}
\crefname{subsection}{Sec.}{Secs.}
\Crefname{subsection}{Section}{Sections}
\Crefname{table}{Table}{Tables}
\crefname{table}{Tab.}{Tabs.}

\spnewtheorem{Def}{Definition}{\bfseries}{\normalfont}
\spnewtheorem{Remark}{Remark}{\bfseries}{\normalfont}
\spnewtheorem{Prop}{Proposition}{\bfseries}{\itshape}
\crefname{Def}{Def.}{Defs.}
\Crefname{Def}{Def.}{Defs.}
\crefname{Prop}{Prop.}{Props.}
\Crefname{Prop}{Prop.}{Props.}
\crefname{Remark}{Remark}{Remarks}
\Crefname{Remark}{Remark}{Remarks}

\titlerunning{Verification of DNNs Using Fuzzy Logic and Concept Embeddings}
\author{
Gesina Schwalbe\inst{1,2}\orcidID{0000-0003-2690-2478}\and
Christian Wirth\inst{1}\and
Ute Schmid\inst{2}\orcidID{0000-0002-1301-0326}
}
\authorrunning{G. Schwalbe et al.}
\institute{
Continental AG, Regensburg, Germany\\
\email{gesina.schwalbe@conti.de}\\
\email{christian.2.wirth@continental-corporation.com}
\and
Cognitive Systems, University of Bamberg, Bamberg, Germany \\
\email{ute.schmid@uni-bamberg.de}}

\begin{document}
\pagestyle{headings}
\mainmatter

\title{Enabling Verification of
Deep Neural Networks in Perception Tasks Using Fuzzy
Logic and Concept Embeddings}

\maketitle

\begin{abstract}
    One major drawback of deep convolutional neural networks (CNNs) for use in safety critical
    applications is their black-box nature. This makes it hard to verify or monitor
    complex, symbolic requirements on already trained computer vision CNNs.
    In this work, we present a simple, yet effective, approach
    to verify that a CNN complies with symbolic predicate logic rules which relate visual concepts.
    It is the first that
    (1)~does not modify the CNN,
    (2)~may use visual concepts that are no CNN in-~or output feature, and
    (3)~can leverage continuous CNN confidence outputs.
    To achieve this, we newly combine methods from explainable artificial intelligence and logic:
    First, using supervised concept embedding analysis, the output of a CNN is post-hoc enriched by concept outputs.
    Second,
    rules from prior knowledge are modelled as truth functions that accept the CNN 
    outputs, and can be evaluated with little computational overhead.
    We here investigate the use of fuzzy logic, i.e., continuous truth values, and of proper output calibration,
    which both theoretically and practically show slight benefits.
    Applicability is demonstrated on state-of-the-art object detectors for three verification use-cases,
    where monitoring of rule breaches can reveal detection errors.
    
\end{abstract}

\begin{figure*}
    \centering
    \vspace*{-\baselineskip}
    \includegraphics[width=.9\linewidth]{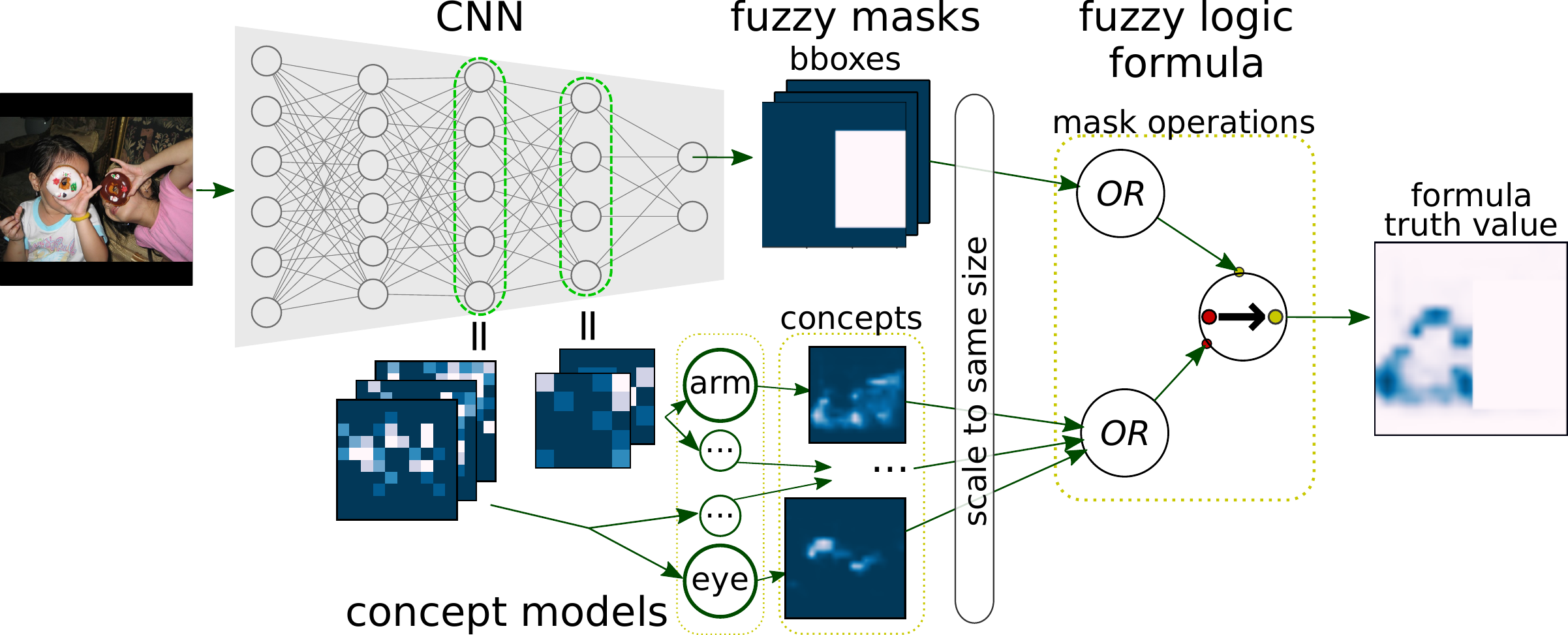}
    \caption{
        Visualization of our approach from \cref{sec:approach} for the pixel-wise formula
        $F(p) = \left( \isPred{eye}(p) \vee \isPred{arm}(p) \vee \dots \right) \rightarrow \isPred{Person}(p)$.
        {\tiny (Image attribution in \cref{fig:cornercases})}.
    }
    \label{fig:overview}
\end{figure*}

\section{Introduction}

Deep neural network (DNN) debugging and safety critical use-cases like automated driving perception
require verification of semantic prior domain knowledge \cite{schwalbe_survey_2020}.
Such knowledge relates concepts that are semantically meaningful to humans, e.g.,
objects, sub-objects, and object properties,
and is potentially complex and fuzzy~\cite{novak_mathematical_1999}
Examples are physical laws or typical anatomical structures.
While it would be desirable to formulate the prior knowledge as verifiable constraints,
this fails for standard black-box object detection CNNs for computer vision tasks \cite{liu_deep_2020}.
In particular, they lack an association of semantic concepts with inputs or (intermediate) outputs:
\begin{enumerate*}[label=(\arabic*), itemjoin={,\ }, itemjoin*={, and\ }]
\item inputs are non-symbolic
\item final outputs only describe few concepts to reduce labeling costs (e.g., few object classes, no sub-object classes)
\item intermediate outputs are non-interpretable
\end{enumerate*}.
Hence, standard work on improving~\cite{donadello_logic_2017} or formally verifying~\cite{liu_algorithms_2021} rule compliance are not applicable without custom-built DNN architectures.
One way to overcome this and post-hoc extract learned concept information from a trained CNN is concept (embedding) analysis (CA)
\cite{bau_network_2017,fong_net2vec_2018,kim_interpretability_2018}.
The idea is to attach cheap and small models to the CNN intermediate outputs.
These are trained in a supervised manner to predict the existence of a concept.
This also works for large object detectors as shown in \cite{schwalbe_verification_2021},
requires only few additional labels \cite{fong_net2vec_2018,kazhdan_disentanglement_2021}, and leaves the original DNN unchanged.

We here combine CA with methods of (fuzzy) formal rule formulation.
This for the first time allows to \emph{post-hoc} score the compliance of CNN outputs \emph{and intermediate outputs} with \emph{predicate logic rules}, such as
formalizations of \enquote{Limbs usually belong to a person}.
For this, the outputs of a CNN are interpreted as family of logical predicates that produce (fuzzy) truth values from image inputs. Rules are modelled to be truth functions accepting these truth values and yielding pixel- or region-wise \emph{logical consistency scores}.
Once the rule model is established, the evaluation works in a \emph{self-supervised} manner, suitable for several debugging and verification use-cases:
\begin{enumerate*}[label=(\arabic*), itemjoin={,\ }, itemjoin*={, and\ }]
    \item
    identification of corner cases (examples with low score) 
    \item
    comparison of the logical consistency of different CNNs 
    \item
    monitoring of logical consistency as error indicator during runtime
\end{enumerate*}.

Besides the mentioned use-cases, this work investigates the benefits of using t-norm \emph{fuzzy logic}~\cite{novak_mathematical_1999} for modelling, and of \emph{calibrating} the newly attached CNN outputs.
Fuzzy logic promises to better leverage the continuity of CNN outputs  \cite{hullermeier_does_2015}.
And, for a mathematically sound interpretation as truth values, CNN outputs should be calibrated (cf.~\cref{sec:approach.calibration}).
Our contributions are:
\begin{enumerate}[nosep, 
                  label=(\roman*), ref={contribution (\roman*)}]
    \item\label{contrib:framework} 
    A novel, simple, extensible and scalable framework
    to construct a \emph{truth value monitor} for logical constraints on semantic concepts
    (\cref{sec:approach}, cf.~\cref{fig:overview}),
    \item\label{contrib:eval.formulations} 
    Development and comparison of \emph{modeling options for formulating the monitor},
    including different fuzzification methods
    (\cref{sec:approach.fuzzylogic,sec:exp.fuzzylogic.formulations}),
    \item\label{contrib:eval.calibrationandfuzziness} Analysis of benefits from \emph{model calibration} and \emph{fuzziness},
    \item\label{contrib:usecases} 
    Demonstration 
    on two \emph{state-of-the-art object detectors}
    and two safety relevant occlusion robustness rules
    for pedestrian detection
    (\cref{sec:exp.fuzzylogic});
    \item\label{contrib:eval.performance} 
    We show that the framework can
    \emph{uncover a substantial proportion of detection errors} with a single rule.
\end{enumerate}

\section{Related Work}
\label{sec:related_work}
\begin{description}[wide=0pt, font=\normalfont\bfseries]
\item[{Fuzzy Logic Integration into DNNs:}]
Diligenti \etal \cite{diligenti_semantic-based_2017} suggested to use fuzzy logic rules as loss functions for DNN training,
relying on continuous t-norm fuzzy logic \cite{novak_mathematical_1999}.
This inclusion of semantic prior knowledge to the training was shown to bring
significant performance improvements for different types of rules
\cite{roychowdhury_image_2018,nandwani_primal_2019,badreddine_logic_2021},
and was extended to several frameworks
\cite{%
    marra_lyrics_2019,
    badreddine_logic_2021
}.
While we build upon this idea,
the prior work requires all referenced concepts to be features in the DNN output, and
does not guarantee that classifier outputs are calibrated.
%
\item[{Concept Embedding Analysis:}] 
The basic works from CA associate semantic concepts with
linear combinations (\emph{concept activation vectors}) of
neurons~\cite{kim_interpretability_2018}  or
CNN filters~\cite{fong_net2vec_2018}.
For this, small linear \emph{concept models} are attached to CNN intermediate outputs.
We use the successor \cite{schwalbe_verification_2021} of \cite{fong_net2vec_2018} as part of our framework (cf.~\cref{sec:approach.ca}),
but apply calibration and optimize the choice of loss towards this.
Prior work proposes two semantic consistency metrics based on CA:
correct similarity between concept encodings~\cite{fong_net2vec_2018,schwalbe_verification_2021},
and attribution of early-layer concepts to later-layer ones
\cite{kim_interpretability_2018,wang_chain_2020}.
Both are restricted in the complexity of verifiable relations,
whereas this work can deal with general predicate logic rules.
\item[{Calibration:}] 
For obtaining calibrated confidence estimates, we use model calibration. 
Common post-calibration methods \cite{guo_calibration_2017} are computationally cheap,
but approximate Bayesian learning is usually outperforming these methods \cite{krishnan_improving_2020}.
However, common variational inference methods \cite{kingma_variational_2015} are mostly
restricted to a mean field approximation, disregarding the covariance,
due to computational cost.
Hence, we use a full covariance Laplace approximation \cite{kristadi_being_2020},
as this is computationally viable with our approach.
Find a comprehensive overview in \cite{abdar_review_2021}.
\item[{Other Monitoring Approaches:}] 
Self-supervised runtime monitoring of DNNs usually relies on uncertainty
estimation, like \cite{rottmann_prediction_2020},
or consistency checks based on pixel-attribution or additional inputs, \forexample LiDAR.
Uncertainty can detect outliers or proximity to decision boundaries \cite{perello-nieto_background_2016}, but no logical inconsistencies.
Pixel-attribution can be used to check for spurious input attribution
patterns of the model \cite{fong_interpretable_2017,lapuschkin_unmasking_2019}, or
for attention not local to detected objects~\cite{cheng_dependability_2018}.
While this is useful for manual inspection, attention methods are
restricted in the rules that can be checked, and
inference of our method requires only a single forward pass.
\end{description}

\section{Approach}\label{sec:approach}

Assume one wants to verify that an object detector respects the rule
\enquote{Heads and limbs belong to a person}
formalized to the predicate logic formula $F$
\begin{align}
    &&F(p) &\coloneqq \Pred{IsBodyPart}(p) \rightarrow \Pred{IsPartOfAPerson}(p)
    \label{eq:fnformula}
    \\
    &\text{with}&
    \Pred{IsPartOfAPerson}(p) &\coloneqq
    \left(
        \exists q\in P\colon \isPred{person}(q) \wedge \Pred{CloseBy}(p, q)
    \right)
    \\
    &\text{and}&
    \Pred{IsBodyPart}(p) &\coloneqq
    {\textstyle\bigvee_{{\text{\Concept{b}}\in\constant{BodyParts}}}} \isPred{b}(p)
\end{align}
with the free variable $p\in\text{Images}\times\text{PixelPositions}$ a pixel position in an image,
and some pre-defined $\constant{BodyParts}$ predicates $\isPred{b}$ for $b \in\constant{BodyParts}$.
Such a formula with free variables can be viewed as a function on the variable values
that outputs a single truth value for \enquote{Is the rule fulfilled}.
It can be modeled as computational tree with the nodes being functions and logical operations, \idest
logical conjunctions $\vee, \wedge, \neg$;
quantifiers $\forall, \exists$;
and predicates like $\isPred{person}$.
The main idea for verifying a rule on CNN outputs is
to interpret the CNN as family of predicates, i.e., functions that output truth values.
We dissect the formula evaluation into the following steps (cf.~\cref{fig:overview}):
\begin{enumerate}[wide=0pt, leftmargin=2ex, labelsep=*]
    \item \textbf{Obtain outputs of predicates} (calibrated, \cref{sec:approach.calibration}) and functions:
        \begin{itemize}[nosep, wide=0pt, leftmargin=2ex, labelsep=*]
        \item Predicates describing a CNN output, like $\isPred{person}$:
        derived from CNN output.
        \item Predicates describing internal knowledge of the CNN,
        like $\isPred{arm}$:
        extracted from the activation maps via concept analysis
        (\cref{sec:approach.ca}).
        \end{itemize}
    \item \textbf{Evaluate the residual computational tree of the formula}
        to obtain a final (fuzzy) truth value.
\end{enumerate}
To leverage the knowledge encoded in the confidences of the DNN outputs,
we also consider to \emph{fuzzify} all logical operations
(\cref{sec:approach.fuzzylogic}).
It should be noted that all components (additional concept outputs, \cref{sec:approach.ca}; calibration, \cref{sec:approach.calibration}; rule evaluation)
only require few standard tensor operations, and thus produce negligible computational overhead.

\subsection{Fuzzy Logic on DNN Outputs}\label{sec:approach.fuzzylogic} 

For fuzzification of logical operations, we rely on the
theory of t-norm fuzzy logic \cite{novak_mathematical_1999}.
We first recapitulate needed basics thereof. 
Then, further modeling aspects for fuzzy rules are detailed, 
and the considered applications. 

\subsubsection{Basics of Fuzzy Logic}\label{sec:approach.fuzzylogic.basics}
Fuzzy logic generalizes standard predicate logic by allowing more than two truth values.
This concerns the logical operations, which are the following:
\emph{Logical connectives} \emph{NOT} ($\neg$), \emph{AND} ($\wedge$), \emph{OR} ($\vee$), and \emph{implication} ($\rightarrow$)
reduce one or several truth values to a single truth value.
\emph{Quantifiers} \emph{all} ($\forall$), and \emph{exists} ($\exists$)
reduce a given domain of values (\forexample pixel positions) to a single truth value
using a body formula.
\emph{Predicates}
take symbol values (constants or instantiated variables), and return a single truth value.
For example, predicates arising from DNN confidence outputs are inherently fuzzy.
Intuitive fuzzy logical connectives may be defined using a t-norm fuzzy logic~\cite{novak_mathematical_1999}.
The t-norm fuzzy logics from \cref{tab:fuzzylogics}
represent a generating basis of all t-norm logics with continuous \emph{AND}
\cite[sec.~2.3.1, p.~48]{novak_mathematical_1999}.
Quantifiers can either be naturally expressed via the chosen logical connectives
by
$\left(\forall x\in X\colon F(x)\right) \coloneqq \left(\bigwedge_{x\in X} F(x)\right)$
and
$\left(\exists x\in X\colon F(x)\right) \coloneqq \left(\bigvee_{x\in X} F(x)\right)$,
or be implemented by a $\mean$ operation, as suggested in \cite{donadello_logic_2017}.
\\
\emph{Notation:} Predicates with binary output are $\binary{\text{underlined}}$.

\begin{table}
    \centering
    \footnotesize
    \caption{
        The fuzzy connectives \emph{AND} (\emph{t-norm}), \emph{OR} (\emph{t-conorm}), \emph{residuated implication} (R-implication) and \emph{strong implication} (S-implication) defined by Boolean logic (Bool) and standard t-norm fuzzy logics \L{}ukasiewicz (\L{}), Goedel/Minimum (G), and Product/Goguen (P).
        Bool applies the standard Boolean operations
        to values binarized at a threshold $\thresh{}=\thresh{Bool}$ (default: 0.5).
    }
    \label{tab:fuzzylogics}
    \begin{tabular}{@{}l c c c c c@{}}
         \toprule
               & Negation & Conjunction & Disjunction & R-Implication      & S-Implication\\
         Logic & $\neg a$ & $a\wedge b$ & $a\vee b$   & $a\rightarrow_R b$ & $(\neg a)\vee b$
         \\\midrule[\heavyrulewidth]
         \L{}
         & $1-a$
         & $\max(0, a + b - 1)$
         & $\min(1, a + b)$
         & $\min(1, 1 - a + b)$
         & $\min(1, 1 - a + b)$
         \\
         G
         & $1-a$
         & $\min(a, b)$
         & $\max(a, b)$
         & $1$ if $a\leq b$ else $b$
         & $\max(1-a, b)$
         \\
         P
         & $1-a$
         & $a \cdot b$
         & $a + b - a\cdot b$
         & $\min(1, \frac{b}{a})$
         & $1 - a + a\cdot b $
        \\
        Bool
        & $\booleanop{\neg} (a\geq \constant{t})$
        & $(a\geq \constant{t}) \booleanop{\wedge} (b\geq \constant{t})$
        & $(a\geq \constant{t}) \booleanop{\vee} (b\geq \constant{t})$
        & $(a < \constant{t}) \booleanop{\vee} (b\geq \constant{t})$
        & $(a < \constant{t}) \booleanop{\vee} (b\geq \constant{t})$
        \\\bottomrule
    \end{tabular}
\end{table}

\subsubsection{Formulation of Rules for Object Detection}\label{sec:approach.fuzzylogic.approach}
Object detection is subject to several fuzzy logic rules
that are based, e.g., on
laws,
physical constraints, or
statistical prior knowledge like typical anatomy.
Simple examples are
spatial rules, e.g.,
\enquote{Road users usually do not fly.}, or
\enquote{A person located on top of a bike is usually a cyclist.};
and hierarchical rules, e.g.,
\enquote{Cars usually have wheels.},
or the example rule from \cref{eq:fnformula}.
Note that the last one describes a safety relevant occlusion robustness:
Infringements of the rule may indicate that the detector
can fail in cases of high occlusion when only few body parts are visible.
To leverage pixel-wise information obtained from the DNN,
we propose a pixel-wise formulation of rules, as done in the example rule from \cref{eq:fnformula}.
This allows to model the logical operations via few standard tensor operations on the mask tensors.
Further details of our modeling approach are given in the following.

For \textbf{fuzzy connectives} we consider the three t-norm fuzzy logics detailed in \cref{tab:fuzzylogics}.
Note that the non-continuous R-implication $a\rightarrow_Rb$ of product and Goedel logic is instable
if truth values of both $a$ and $b$ are small~\cite{badreddine_logic_2021}.
As baseline we consider non-fuzzy Boolean logic, where $\wedge$ and $\vee$ translate to standard mask intersection and union.
For this, all inputs to logical connectives must be binarized at a threshold $\thresh{Bool}$ (default 0.5).
This may reduce memory consumption, but discards potentially valuable confidence information.

In \cite{donadello_logic_2017}, the \textbf{fuzzy quantifier} definition $\forall=\mean$ was suggested 
as an intuitive and differentiable implementation.
Note that, other than $\forall$ quantifiers derived from a fuzzy \emph{OR},
$\mean$ gives rise to the following unequality
for a subset $Q\subset P$ and a t-norm fuzzy implication:
\begin{align}
    \left(\forall p\in Q\colon F(p)\right)
    \label{eq:meanforall1}
    \neq &\left(\forall p\in P\colon (p\in Q) \rightarrow F(p)\right)
\end{align}
The right hand formulation produces generally large values for small $Q$\footnote{
    $\left(\forall p\in P\colon (p\in Q) \rightarrow F(p)\right)
    = \tfrac{\#P - \#Q}{\#P} + \tfrac{\#Q}{\#P}\mean_{p\in Q}F(p)$
}.
Thus, to keep intuition intact, we recommend the left hand formulation from \eqref{eq:meanforall1} where applicable.
When choosing the $\exists$ quantifier, it must be noted that for $\forall=\mean$ the standard definition 
$\left(\exists x\colon F(x)\right) = \left(\neg \forall x\colon \neg F(x)\right)$
for a formula body $F$ would produce $\exists=\mean$.
This ignores high truth values when their proportion in the domain is too low.
Thus, a $\exists$ defined from a fuzzy logic $\bigwedge$ is preferrable.

Typical \textbf{predicates} occurring in rules for object detection are
unary concept predicates (\enquote{Does the concept apply to the location/region?})
and spatial relations.
Using examples from \cref{eq:fnformula}, we demonstrate different ways to model predicates:
using predicted or ground truth (GT) bounding boxes (e.g., $\isPred{person}, \isPred{GTPerson}$),
using concept segmentation masks (e.g., $\Pred{IsBodyPart}$),
and manually (e.g., $\Pred{CloseBy}$).

\begin{compactdescription}[font=\normalfont\itshape, topsep=0pt]
\item[Box information.]
    Instance-wise bounding box information for an object class
    can be turned into a fuzzy semantic segmentation mask, e.g., with the following simplistic transformation:
    \begin{enumerate*}[label=(\arabic*), itemjoin={,\ }, itemjoin*={, and\ }]
        \item Each bounding box is turned into a mask by setting pixels inside the bounding box to their bounding box objectness score value
        \item if needed, the box masks are combined using pixel-wise logical \emph{OR} of the respective fuzzy logic
    \end{enumerate*}. 
    For GT bounding boxes, the objectness score may be set constant to 1.
\item[Segmentation masks.]
    Segmentation outputs, \forexample from CA (here called \emph{concept masks}, cf.~\cref{sec:approach.ca}),
    can serve as unary concept predicates, each pixel value being a truth value for a pixel position.
    The masks of different concepts may differ in resolution,
    \forexample those from CA have the resolution of CNN activation maps.
    To ensure compatible sizes, one can:
    \begin{enumerate*}[label=(\alph*), itemjoin*={, or\ }]
    \item upscale masks to the common largest resolution, \forexample bilinearly~\cite{schwalbe_verification_2021}
    \item downscale masks to the common smallest resolution, \forexample using maxpooling%
        \footnote{
            Formally, maxpooling on a unary predicate $\isPred{X}$ can be written as
            $\exists p'\in P'\colon \Pred{CloseBy}(p, p')\wedge \isPred{X}(p')$
            for a pixel $p\in P$ on some other resolution $P'$, and binary
            $\Pred{CloseBy}(p, p')=(\|p-p'\|_1\leq \lfloor\frac{1}{2}(\constant{ksize}-1\rfloor))$
            (cf.~\cref{eq:neighborcond2}).
        }.
    \end{enumerate*}

\item[$\Pred{CloseBy}$.]
    We suggest two formulations of a spatial close-by relations on pixel coordinates $a, b$,
    a Gaussian radial distance function (with a low cut for memory efficiency),
    and a simpler binary one that checks $L_1$ proximity:
    \begin{align}
        \SwapAboveDisplaySkip
        \Pred{CloseBy}_{\sigma, r}(a, b) &\coloneqq \exp(- d(a,b)^2 / (2\sigma^2) )\;
        \label{eq:closeby}
        \\
        \Pred{CloseBy}(p, q) &\coloneqq (\|p-q\|_1 \leq \lfloor\tfrac{1}{2}(\constant{ksize}-1)\rfloor)\;,
        \label{eq:closebykernel}
    \end{align}
    with $d(a,b) \coloneqq \|a-b\|_2$ if $\|a-b\|_1 \leq r$
    for a window half size $r$, else 0.
    $\Pred{CloseBy}_\sigma$ becomes trivial if $\sigma=0$, i.e.,
    $\Pred{CloseBy}_{0}(a, b) = 1$ if $a\equiv b$ else $0$.
    Note that $\Pred{CloseBy}$ can be defined from a neighborhood definition $\Pred{Nbh}(\cdot)$ via set membership: $\Pred{CloseBy}(p, q) = (q \in \Pred{Nbh}(p))$ (cf.~\cref{eq:meanforall1}).
    For \cref{eq:closebykernel} $\Pred{Nbh}(p)$ is a square window of size $\constant{ksize}$ around $p$.
    A non-trivial spatial relation may help to mitigate segmentation mask resolution problems:
    Upscaling low resolution masks may cause positive regions to become too large.
    For example, in the rule from \cref{eq:fnformula} body part masks may \enquote{shoot over} the person areas,
    leading to rule infringements at person area boundaries (cf.~\cref{fig:cornercases}).
\end{compactdescription}

One \textbf{pattern} for rule formulation we consider here is that of a neighborhood prior for pixel-wise formulas $F$.
The neighbor condition $\Pred{NbCond}_F(p, P)$ should down-weight $F(p)$ if it is spatially isolated (noise) within the region $P$.
We suggest to score $p$ isolated if
none (\cref{eq:neighborcond2}) or, alternatively, not all (\cref{eq:neighborcond1}) of
$p$'s neighbors have a high $F(p)$ value, formally:
\begin{align}
    \Pred{NbCond}_F(p, P) &= \forall q\in P\colon 
        \Pred{CloseBy}(p, q) \rightarrow F(q)
        \overset{\text{\text{\tiny binary}}}{\underset{\mathclap{\Pred{CloseBy}}}{=}}
        \forall q\in \Pred{Nbh}(p)\colon M(q)
        \label{eq:neighborcond1}\\
    \Pred{NbCond}_F(p, P) &= F(p) \wedge \left(
        \exists q\in P\setminus\{p\}\colon \Pred{CloseBy}(p, q) \wedge F(q)\right)
        \label{eq:neighborcond2}
\end{align}
If $\Pred{CloseBy}$ is radial symmetric, \cref{eq:neighborcond1} selects center points of radial peaks.
For $\Pred{CloseBy}$ defined via a square $\Pred{Nbh}$ as in \cref{eq:closebykernel},
and using \cref{eq:meanforall1} and $\forall=\mean$, the neighborhood condition
\cref{eq:neighborcond1} can be efficiently implemented as a stride~1 average pooling.
In this case, to not loose any peak, the kernel size $\constant{ksize}$ is best chosen to cover at least the minimum expected area of an interesting peak.

\subsubsection{Fuzzy Rules for Verification}\label{sec:approach.fuzzylogic.applications}
Assume one is given a logical rule of the form
$R(P)=\forall p\in P\colon F(p)$
where the variable $P$ is an image or image region, and $p\in P$ means a pixel position in that region, and $F$ a truth function on such valid pixels $p\in\text{ImageRegions}\times\text{PixelPositions}$, like in \cref{eq:fnformula}.
For a concrete image, the pixel outputs of $F$ together form a mask with values in $[0,1]$ that are high where the formula is fulfilled, and low otherwise.
$R$ yields for an image or image region $P$ a single truth value that can be interpreted as a logical consistency score for $P$.
We suggest the following use-cases to utilize $R$ and $F$ for verification, given a test set $T$.
\begin{description}[wide=0pt, font=\normalfont\itshape, topsep=0pt]
    \item[Corner Case Search:]
        $R$ can be evaluated for each image in $T$.
        Outliers with respect to the consistency score distribution may serve as pre-selection for manual analysis
        in order to identify root causes of illogical behavior.
    \item[Global Logical Consistency Score:]
        The aggregation $\forall P\in T\colon  R(P)$ of image-wise scores over $T$ yields a global consistency score over $T$.
        The score allows, e.g., comparison of different networks trained
        on the same dataset, or comparison of different training datasets.
    \item[Monitoring:]
        The pixel-wise truth scores given by $F$ give rise to several runtime monitoring scenarios.
        For our example formula \cref{eq:fnformula}, a monitor would have the task
        to reveal cases of detector false negatives where some body part was found
        but no person predicted in the end.
        A \emph{pixel-wise monitor} that observes whether pixels $p$ have a suspiciously low
        truth value can be defined as
        \begin{gather}
            M(p) \coloneqq \neg F(p) \in [0, 1].
            \label{eq:pixelmonitor}
        \end{gather}
        Alarm is raised if an inconsistency bound is breached, i.e., $M(p) \geq \thresh{px}$.
        The benefit of the pixel-wise approach is that errors can be localized.
        Hence, only suspicious parts of the prediction can be discarded;
        or warnings can ignored for image regions that are not of interest,
        such as regions far off the vehicle trajectory
        in case of a pedestrian detector.
        A \emph{region-wise monitor} for image regions $P$ marks complete regions or images as spurious based on the pixel truth values.
        We suggest two definitions derived from $M$, a simple one and a smoothed one:
        \begin{align}
            \SwapAboveDisplaySkip
            M_{\text{reg}}^{\text{simple}}(P) &\coloneqq \exists p\in P\colon M(p)
            \overset{{(*)}}{=} \neg R(P)
            \label{eq:monitoregionsimple}
            \\
            M_{\text{reg}}^{\text{peaks}}(P) &\coloneqq \exists p\in P\colon \Pred{NbCond}_M(p, P)
            \overset{{(**)}}{=}
            \max_{p\in P} \text{AvgPool2D}\left(\big(M(p)\big)_{p\in P}\right)
            \label{eq:regionmonitoradvancedours}
        \end{align}
        where equality $(*)$ holds if
        $(\exists x\colon f(x)) \equiv (\neg\forall x\colon\neg f(x))$,
        and equality $(**)$ for $\exists=\max$ and the average pool formulation of $\Pred{NbCond}$ from \cref{eq:neighborcond1}.
        The smoothing in \cref{eq:regionmonitoradvancedours} essentially adds a prior on the shape of peaks of monitor pixel values.
        In general, the choice of quantifier is important:
        Some are sensitive to outliers, \idest isolated pixel alarms, and need the smoothing in \cref{eq:regionmonitoradvancedours} (\forexample $\exists=\max$ from Goedel logic),
        others are sensitive to the ratio between alarm and non-alarm pixels (\forexample $\mean$),
        or are computationally expensive (\forexample $\exists=\bigvee$ for the probabilistic sum $\bigvee$ from Product logic).
\end{description}

\subsection{Concept Analysis for Additional DNN Outputs}\label{sec:approach.ca} 

Standard object detection outputs usually include only a small set of object classes,
\forexample \Concept{person},
and no sub-objects or other object attributes allowing for rich semantics.
Our goal is to post-hoc enrich the output of a DNN by those concepts that are necessary to
formulate given prior domain knowledge like \cref{eq:fnformula} on DNN inputs and outputs.
This should be done
\begin{enumerate*}[label=(\arabic*), itemjoin={,\ }, itemjoin*={, and\ }]
\item without changing the trained DNN
\item with little additional training and labeling effort
\item extensible, \idest allowing to later add further concepts in case more domain knowledge is collected
\end{enumerate*}.
For this, supervised post-hoc concept analysis constitutes a suitable candidate.
\emph{Concept (embedding) analysis} (CA) in general aims to associate in a simple way semantic concepts with
elements in the intermediate output of a DNN, usually that of a layer \cite{schwalbe_verification_2021}.
Standard types of visual semantic concepts are texture, material, objects, object parts, or image-level concepts like the scene \cite{bau_network_2017}.
To achieve the association, supervised CA methods learn to predict information
about the concept of interest, \forexample a binary segmentation mask, from the
intermediate outputs.
We here use the CA method suggested in \cite{schwalbe_verification_2021},
which was shown to work on CNN object detection.
The concept models are attached to the output of one CNN layer.
Each consists of a
$1\times1$-convolution, followed for inference by a sigmoid normalization (and potentially upscaling).
This method only introduces minor computational overhead.
The output of the concept models are pixel-masks with confidence values in $[0,1]$,
which are trained to match ground truth binary segmentations of the concept.
Due to our additional need for well-calibrated outputs (\cref{sec:approach.calibration}),
we suggest to exchange the original self-balancing losses from \cite{schwalbe_verification_2021} for standard binary cross-entropy loss (BCE) which has less impact on calibration.

\subsection{Concept Model Calibration}\label{sec:approach.calibration}
DNN confidence outputs, including our post-hoc attached concept models,
may be badly calibrated.
This means that the confidence values do not well match
the actual probability with which the prediction is correct \cite{guo_calibration_2017}.
When interpreting confidences as fuzzy truth values, bad calibration violates
the fundamental assumption that the truth \emph{meaning} monotonously
increases with the truth \emph{value}.

Therefore, we suggest to employ approximate Bayesian learning for calibrating our concept models (cf.~\cref{sec:related_work}).
We compute a full covariance approximation of the posterior via Laplace approximate \cite{kristadi_being_2020}. Despite the required Hessian approximation, we still maintain limited computational overhead, as we only apply it to a single layer (cf.~\cref{sec:approach.ca}) with few parameters. This method can be applied without changing or retraining the base model, making it applicable to any pretrained network. 


\section{Experiments}\label{sec:exp}

\begin{figure*}[t]
    \includegraphics[width=.499\linewidth]{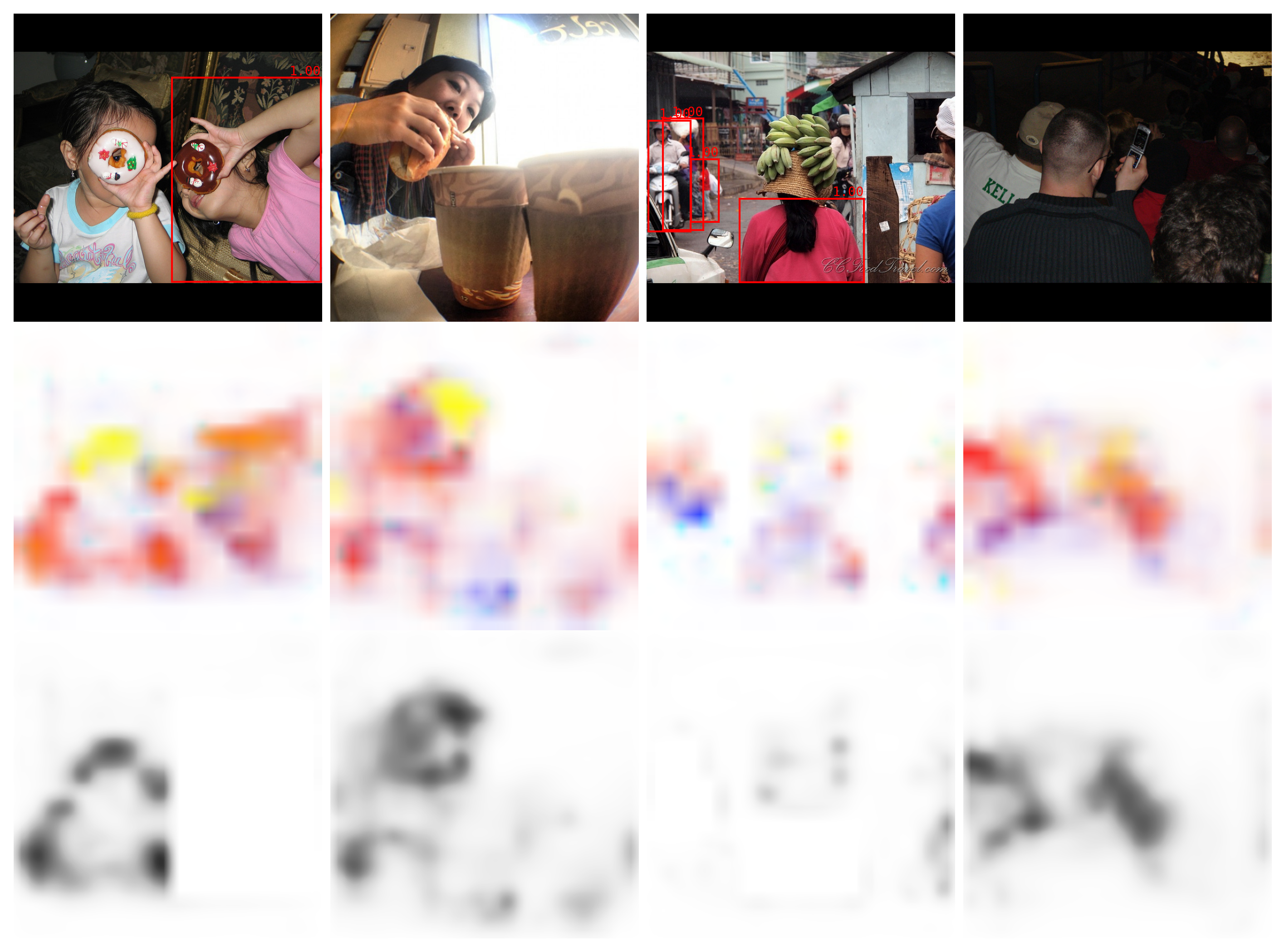}%
    \hfill%
    \includegraphics[width=.499\linewidth]{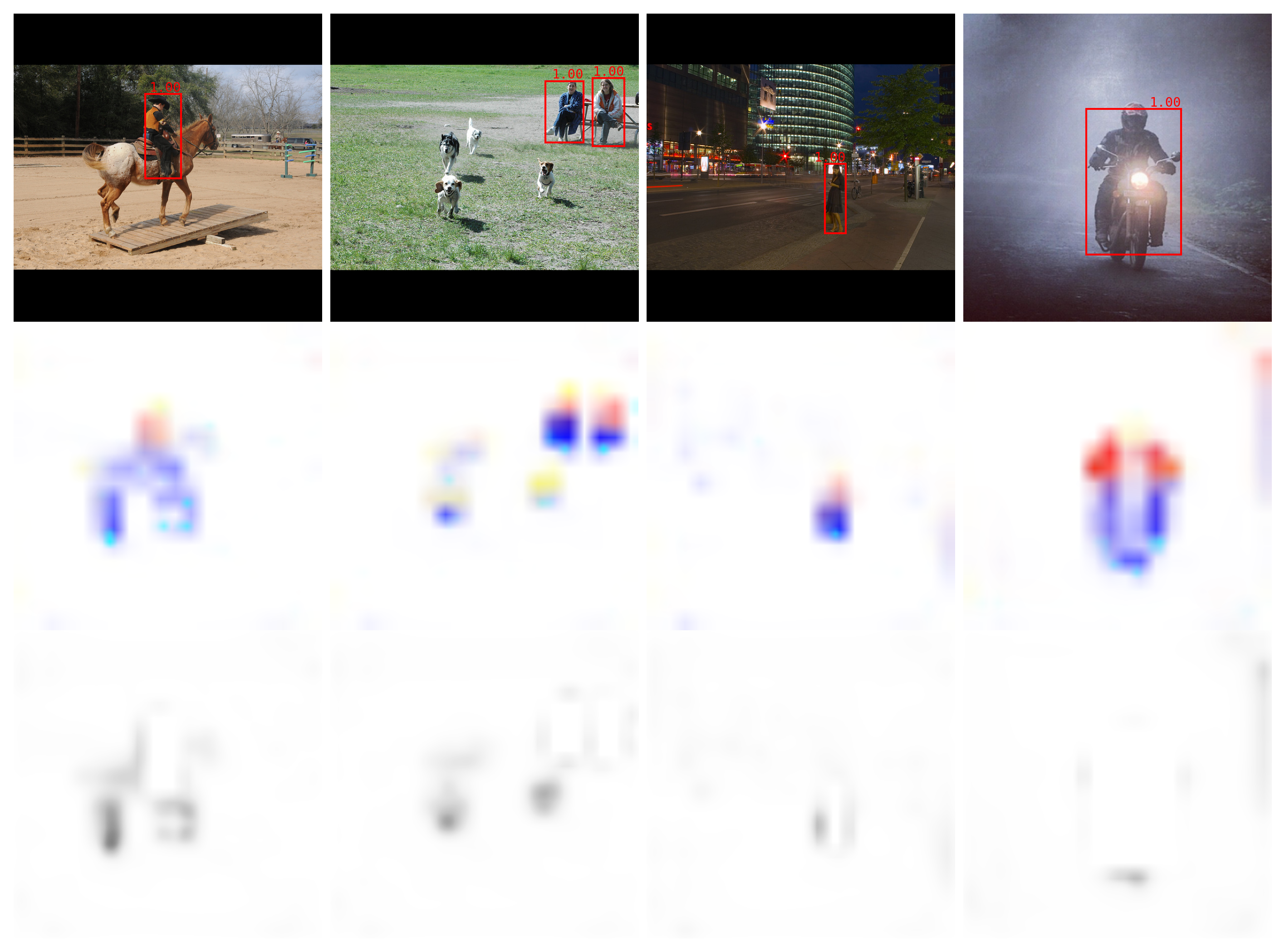}%
    {\tiny\par
        \noindent Left to right thanks to
        \href{http://flickr.com/photo.gne?id=2372745068}{David Quitoriano},
        \href{http://flickr.com/photo.gne?id=8239498722}{j bizzie},
        \href{http://flickr.com/photo.gne?id=8026451116}{CCFoodTravel.com},
        \href{http://flickr.com/photo.gne?id=3228843052}{k.steudel},
        \href{http://flickr.com/photo.gne?id=8653189194}{Fort Rucker},
        \href{http://flickr.com/photo.gne?id=2516944023}{Donelle},
        \href{http://flickr.com/photo.gne?id=190818272}{Jacob Bøtter},
        \href{http://flickr.com/photo.gne?id=7243537732}{Vir Nakai};
        \textcopyright~\href{http://creativecommons.org/licenses/by/2.0/}{CC~BY~2.0}
        \par\vspace*{-\baselineskip}
    }
    \caption{
    EffDet predicted boxes (\emph{top}),
    concepts ({\small%
    {\color[HTML]{0000FF}\Concept{leg}},
    {\color[HTML]{00FFFF}\Concept{ankle}},
    {\color[HTML]{FF0000}\Concept{arm}},
    {\color[HTML]{FFA500}\Concept{wrist}},
    {\color[HTML]{FFFF00}\Concept{eye}}};
    \emph{middle row}), and
    logical consistency pixel values (\emph{bottom})
    for \cref{eq:regionmonitoradvancedours} (before $\max$) with P logic cal setting from \cref{tab:auc}.
    \emph{Left}: Monitor true positives; \emph{right}: false positives.
    }
    \label{fig:cornercases}
\end{figure*}

This section discusses evaluation results for
concept model creation (\cref{sec:exp.ca}) and
the three verification use-cases (\cref{sec:exp.fuzzylogic}), as well as
limitations of our approach (\cref{sec:exp.limitations}).
%
The use-cases are tested on the example occlusion robustness rule from \cref{eq:fnformula},
and for body parts
\Concept{eye}, \Concept{arm}, \Concept{wrist}, \Concept{leg}, \Concept{ankle}.
%
The $\isPred{person}$ predicate is derived from \Concept{person} bounding box outputs of the
two networks
Mask R-CNN~\cite{he_mask_2017} (MR) with ResNet50 backbone,
and EfficientDet~D1~\cite{tan_efficientdet_2020} (EffDet).
The weights are trained on the MS~COCO train2017 dataset~\cite{lin_microsoft_2014},
and are taken from PyTorch~\cite{pytorch_torchvision_2021} respectively TensorFlow modelzoo~\cite{wightman_efficientdet_2021}
(cf.~\cref{tab:modelresults} for statistics).
%
The ground truth for the body part concepts was derived
from the MS~COCO 2017~\cite{lin_microsoft_2014} keypoint annotations
as done in \cite{schwalbe_verification_2021}.
Concept models were trained and calibrated on the MS COCO train2017 dataset,
with a train/validation split of 4:1.
Evaluation results both for the concept model performance and the fuzzy logic
are collected on the MS COCO val2017 dataset.
The performance of $\isPred{person}$ and the concept model segmentations
is evaluated as the set intersection over union (sIoU)~\cite{fong_net2vec_2018}
between the binary ground truth concept masks $(m_i)_i$ and
the predicted and upscaled concept masks $(m_i^{\text{pr}})_i$:
\begin{gather}
  \text{sIoU}\left((m_i)_i, (m_i^{\text{pr}})_i\right)
  = \tfrac
  {\sum_i \sum m_i \cap (m_i^\text{pr}>\thresh{sIoU})}
  {\sum_i \sum m_i \cup (m_i^\text{pr}>0.5)}\;.
\end{gather}
\emph{Notation:} \tth{sIoU} is measured at optimal $\thresh{sIoU}$
(determined on a the validation set), sIoU at 0.5.
%

\begin{table}[t]
    \centering
    \caption{Statistics of used models (\ref{tab:modelresults}, \ref{tab:caresults.perconcept}) and global consistency scores (\ref{tab:globalscore})}
    \begin{minipage}[t]{.48\linewidth}
    \begin{subtable}[t]{\linewidth}
    \centering
    \caption{
        Statistics of used $\isPred{person}$ predicates
    }
    \label{tab:modelresults}
    \begin{tabular}{@{}l
    S[table-format=1.3] S[table-format=1.3] S[table-format=1.3]
    @{}}
        \toprule
        {Model}
          & {ECE}    & {Pixel-Acc.} & {sIoU}
          \\
        \midrule[\heavyrulewidth]
        MR~\cite{he_mask_2017} 
          & 0.044096 & 0.955904     & 0.810289  
          \\
        EffDet~\cite{tan_efficientdet_2020}
          & 0.070773 & 0.929227     & 0.683913  
          \\
        \bottomrule
    \end{tabular}
    \end{subtable}\\%
    \begin{subtable}{\linewidth}
        \centering
        \caption{
            Global logical consistency scores
            for rule from \cref{eq:fnformula}
        }
        \label{tab:globalscore}
        \newcolumntype{F}{S[table-format=1.3]}
        \begin{tabular}{@{}lFFFFFF@{}}
        \toprule
        & {G}      & {G cal}  & {\L}    & {\L{} cal}   & P     & {P cal} \\
        \midrule[\heavyrulewidth]
        MR
        & 0.994437 & 0.994099
        & 0.991583 & 0.991033
        & 0.99187  & 0.991344
        \\
        EffDet
        & 0.98784  & 0.98514
        & 0.980968 & 0.975453
        & 0.981759 & 0.976553
        \\
        \bottomrule
        \end{tabular}
    \end{subtable}
    \end{minipage}%
    \hfill%
    \begin{subtable}[t]{.48\linewidth}
        \caption{
            Layer (L) and performance of the used BCE-trained concept models.
        }
        \label{tab:caresults.perconcept}
        \begin{tabular}{@{}l 
        @{~}c@{~~} S[table-format=1.3] 
        @{~~} S[table-format=1.3]
        @{\hspace{2ex}}
        c@{~~} S[table-format=1.3] 
        @{~~} S[table-format=1.3] 
        @{}}
            \toprule
            &\multicolumn{3}{c}{\bfs MR}
            &\multicolumn{3}{c}{EffDet}
            \\
            {C}             &{L}& {\tth{sIoU}}  & {sIoU}  
                            &{L}& {\tth{sIoU}}  & {sIoU}  \\
            \midrule[\heavyrulewidth]
            \Concept{ankle} & 3 &\bfs0.209613  & 0.132022 
                            & 4 &    0.142438  & 0.058939 \\
            \Concept{arm}   & 4 &\bfs0.329739  & 0.254347 
                            & 6 &    0.304920  & 0.242458 \\
            \Concept{eye}   & 3	&\bfs0.436298  & 0.424009 
                            & 5	&    0.340421  & 0.295704 \\
            \Concept{leg}   & 3	&\bfs0.325850  & 0.264188 
                            & 5	&    0.310094  & 0.246176 \\
            \Concept{wrist} & 4 &\bfs0.184454  & 0.118733 
                            & 6 &    0.149572  & 0.068726 \\
            \bottomrule
        \end{tabular}%
    \end{subtable}%
\end{table}

\begin{table}[t]
    \centering
    \small
    \caption{
        Calibration and performance of best concept models for MR
        (statistics averaged over concepts)
    }
    \label{tab:caresults}
        \small
        \sisetup{round-precision=3}
        \newcolumntype{F}{S[separate-uncertainty,table-format=1.3(1)]}
        \begin{tabular}{@{}lFFFS[separate-uncertainty,table-format=1.2(1)]F@{}}
            \toprule
                       &    {ECE}       &    {MCE}     &  {\tth{sIoU}}       & {$\thresh{\tth{sIoU}}$}            & {sIoU}\\
            \midrule[\heavyrulewidth]
        \bfs     BCE   &    0.001+-0.000 &    0.146+-0.079 &    0.297+-0.102 & 0.23+-0.06 &    0.218+-0.126 \\
        \bfs\hfill cal &\bfs0.001+-0.000 &\bfs0.140+-0.082 &\bfs0.297+-0.102 & 0.24+-0.06 &    0.218+-0.126 \\
            \midrule
            Dice       &    0.010+-0.008 &    0.621+-0.082 &    0.265+-0.079 & 0.59+-0.20 &    0.263+-0.081 \\
            \hfill cal &    0.001+-0.001 &    0.451+-0.121 &    0.264+-0.080 & 0.52+-0.07 &\bfs0.263+-0.080 \\
            \midrule
            bBCE       &    0.083+-0.047 &    0.879+-0.037 &    0.115+-0.052 & 0.95+-0.00 &    0.047+-0.031 \\
            \hfill cal &    0.022+-0.012 &    0.772+-0.071 &    0.228+-0.054 & 0.91+-0.05 &    0.056+-0.032 \\
            \bottomrule
        \end{tabular}
\end{table}

\begin{table*}[t]
    \centering
    \caption{
        Performance comparison of different formula formulations
        for image-level monitoring of rule \cref{eq:fnformula}.
        Metrics: precision-recall (PR) curve, area under ROC curve (AUC),
        F1 at $\thresh{reg}^\text{peaks}=0.5$, and
        each \ol{F1}, \ol{F0.1}, and \ol{F10} for the $\thresh{reg}^\text{peaks}$ yielding the theoretically best corresponding score value.
    }
    \label{tab:auc}
    \begin{subtable}[t]{.5\linewidth}
        \caption{Image-level results for formula \cref{eq:fnformula}}
        \newcolumntype{T}{S[table-format=1.3,round-mode=places,round-precision=3]}
        \newcolumntype{F}{S[table-format=1.3,round-mode=places,round-precision=3,detect-weight=false]}
        \newcommand{\thr}[2][best]{$\text{#2}_{\mathrlap{\text{#1}}}$}
        \small
        \begin{tabular}{@{}>{\small}l@{~}l@{~}TTTTT @{}}
        \toprule
        &           & {AUC}       & {\thr[0.5]{F1}}&{\ol{F1}}& {\ol{F0.1}}& {\ol{F10}}\\
        \midrule[\heavyrulewidth]
        \multirow{6}{*}{\rotatebox{90}{Mask R-CNN}}
        &{Bool}     &    0.618611 &    0.145535 &    0.455094 &    0.421375 &    0.974931  \\
        &{Bool cal} &    0.619923 &    0.143646 &    0.458085 &    0.422991 &\bfs0.975022\\
        &{\L}       &    0.631521 &    0.281915 &    0.461905 &    0.483785 &    0.974946\\
        &{\L{} cal} &    0.631790 &\bfs0.295197 &    0.460905 &\bfs0.492369 &    0.974856\\
        &{P}        &    0.632027 &    0.243665 &\bfs0.466340 &    0.466985	&    0.974946\\
        &{P cal}    &\bfs0.632498 &    0.242718 &    0.464974 &    0.469792	&    0.974856\\
        \midrule
        \multirow{6}{*}{\rotatebox{90}{EfficientDet D1}}
        &{Bool}     &    0.671179 &    0.420806 &    0.749186 &    0.898743	&    0.993404\\
        &{Bool cal} &    0.675827 &    0.410571 &    0.749071 &    0.895071	&    0.993410\\
        &{\L}       &    0.690377 &    0.557311 &    0.751780 &    0.894969	&    0.993410\\
        &{\L{} cal} &    0.694892 &\bfs0.591702 &\bfs0.754563 &\bfs0.899394	&    0.993410\\
        &{P}        &    0.689174 &    0.500000 &    0.751247 &    0.897335	&    0.993410\\
        &{P cal}    &\bfs0.696107 &    0.522920 &    0.753497 &    0.895855	&    0.993410\\
        \bottomrule
        \end{tabular}%
    \end{subtable}
    \hfill%
    \begin{subtable}[t]{0.45\linewidth}
        \centering
        \caption{Upper left part of PR curves}
        \vspace*{-.5\baselineskip}
        \includegraphics[width=.75\linewidth]{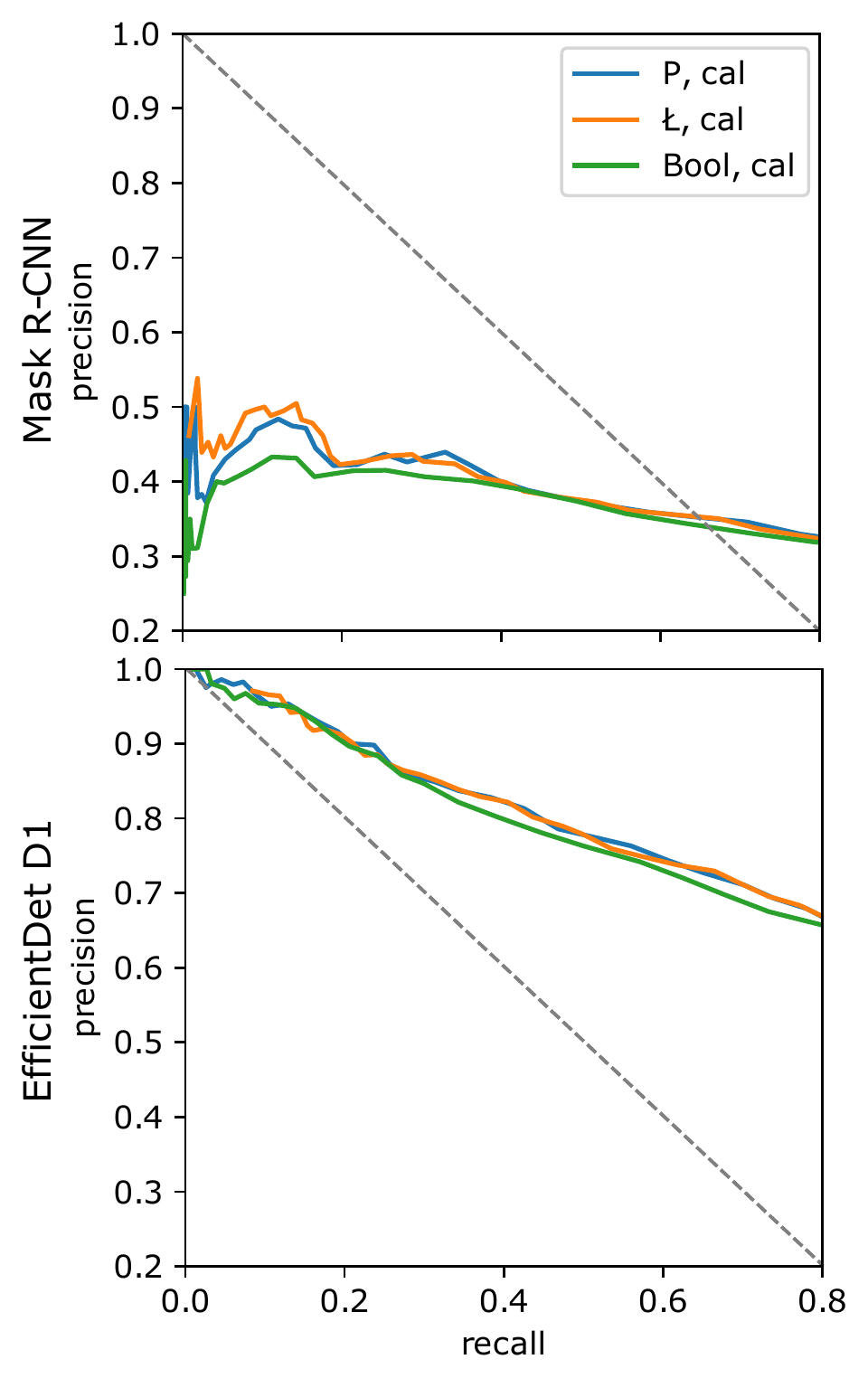}
        \vspace*{-1.5\baselineskip}~
    \end{subtable}
\end{table*}

\begin{table}[t]
    \centering
    \caption{Performance of monitors by binarization threshold $\thresh{px}$, resp. $\thresh{reg}$.
    }
    \label{tab:perpxandperpredresults}
    \newcolumntype{T}{S[table-format=1.3,round-mode=places,round-precision=3]}
    \begin{subtable}[t]{0.3\linewidth}
        \centering
        \caption{Pixel-level ROC AUC for rule from \cref{eq:fnformula}}
        \label{tab:perpixelauc}
        \begin{tabular}[t]{@{}l@{~\,} TTT@{}}
            \toprule
                         & {MR}     & {EffDet}
            \\\midrule[\heavyrulewidth]
               {Bool}    & 0.828949 & 0.839651
             \\{Bool cal}& 0.829743 & 0.843178
             \\{\L}      & 0.832856 & 0.842595
             \\{\L{} cal}& 0.833463 & \bfs 0.847183
             \\{P}       & 0.833019 & 0.842530
             \\{P cal}   & \bfs 0.833626 & 0.847085
             \\\bottomrule
        \end{tabular}
    \end{subtable}%
    \hfill%
    \begin{subtable}[t]{.6\linewidth}
        \centering
        \caption{Prediction-level performance for rule \cref{eq:fpformula}}
        \label{tab:resultsfprule}
    \newcommand{\thr}[2][best]{$\text{#2}_{\mathrlap{\text{#1}}}$}
    \begin{tabular}{@{}>{\small}l@{~}l@{~}TTTTT @{}}
    \toprule
    &           & {AUC}       & {\thr[0.5]{F1}}&{\ol{F1}}& {\ol{F0.1}}& {\ol{F10}}\\
    \midrule[\heavyrulewidth]
    \multirow{3}{*}{\rotatebox{90}{MR}}
    &{Bool cal} &    0.928946 &    0.123450 &    0.230396 &    0.145452 &    0.854433\\
    &{\L{} cal} &    0.927052 &\bfs0.135945 &\bfs0.250538 &\bfs0.172029 &    0.843659\\
    &{P cal}    &\bfs0.930142 &    0.131856 &    0.250267 &    0.171884	&\bfs0.847185\\
    \midrule
    \multirow{3}{*}{\rotatebox{90}{EffDet}}
    &{Bool cal} &\bfs0.986341 &    0.023644 &\bfs0.117647 &\bfs0.115429	&\bfs0.479696\\
    &{\L{} cal} &    0.889197 &\bfs0.027641 &    0.081633 &    0.049571 &    0.463803\\
    &{P cal}    &    0.984955 &    0.027515 &    0.080000 &    0.048383	&    0.452318\\
    \bottomrule
    \end{tabular}
    \end{subtable}%
    \\[-\baselineskip]
\end{table}


\subsection{Concept Analysis and Calibration}\label{sec:exp.ca}
Literature suggests several losses for concept analysis, including
balanced Dice loss \cite{schwalbe_verification_2021,rabold_expressive_2020}, and
a class-balanced binary cross-entropy (bBCE) \cite{fong_net2vec_2018}.
We here also consider standard BCE.
\cref{tab:caresults} compares calibration and performance of these losses in terms of
sIoU, expected calibration error (ECE) and maximum calibration error (MCE)~\cite{guo_calibration_2017}.
Results show that Laplace-calibrated BCE outperforms 
all other variants, \emph{given a tuned sIoU threshold}.
Interestingly, balanced losses (Dice, bBCE) suffer from substantial miscalibration.
This can be partially countered with the Laplace method, but results are still worse. Calibrating the BCE result only shows minor advantages, therefore we re-evaluated the calibration effects for the full approach (cf.~\cref{tab:auc}). 
Improvements are still small, but consistent (cf.~\cref{sec:exp.fuzzylogic.formulations,sec:exp.fuzzylogic.monitor}).
The sIoU performance of the BCE-trained concept models used in later experiments
is given in \cref{tab:caresults.perconcept}.
Representations of the larger Mask~R-CNN model consistently outperform those of EfficientDet~D1 (cf.~\cref{sec:exp.fuzzylogic.globalconsistency}).

\subsection{Logical Consistency Monitor Applications}\label{sec:exp.fuzzylogic}

If not stated otherwise (cf.~\cref{sec:exp.fuzzylogic.formulations}),
experiments were conducted for the S-implication formulation of the formula from \cref{eq:fnformula},
with trivial $\Pred{CloseBy}_\sigma$ from \cref{eq:closeby},
and the arithmetic mean for $\forall$, and $\max$ for $\exists$
(this is $\bigvee_x F(x)$ in Goedel logic).
The Goedel exists quantifier was chosen as it is the most conservative one of those defined by fuzzy logics, \idest yields the smallest values \cite[Lemma~2.19]{novak_mathematical_1999}.
This is a desirable property for safety evaluation with the example rule. 
Concept masks are bilinearly upscaled to ensure comparability amongst CNNs.
For non-fuzzy logic (\cref{tab:fuzzylogics}), the concept masks are binarized at
the same thresholds as the monitor outputs,
\idest $\thresh{Bool}=\thresh{px},\thresh{reg}^\cdot$
for $\thresh{px},\thresh{reg}^\cdot$ defined below.
%
A pixel $p$ is
a \emph{false negative pixel} ($\binary{\isPred{FN}}(p)$), if
$\isPred{FN}(p) \geq \thresh{\Concept{ped}}=0.5$ for
$\isPred{FN}(p) \coloneqq \neg\isPred{person}(p) \wedge \isPred{GTPerson}(p)$.
Since the logical consistency monitor \cref{eq:fnformula} shall
highlight false negatives of the detector, $\binary{\isPred{FN}}$ is taken as pixel-wise ground truth for the monitor.
%
For evaluation of the pixel-level monitor we binarize the outputs of $M$ (\cref{eq:pixelmonitor})
at threshold $\thresh{px}$.
Note that our formula will only highlight those parts of false negative areas
for which the selected body parts were predicted (\forexample not the torso).
Hence, pixel-level recall is naturally comparatively low.
%
To investigate suitability for finding images with \emph{some}
logical inconsistency, we have a look at the region monitor formulations
$M_\text{reg}^\text{simple}$ (\cref{eq:monitoregionsimple}) and
$M_\text{reg}^\text{peaks}$ (\cref{eq:regionmonitoradvancedours})
for complete images as regions.
The ground truth $\Pred{GT}_\text{reg}^\cdot(P)$ for an image $P$ is defined from the pixel-wise
$\binary{\isPred{FN}}$ values using the same formula as for deriving $M_\text{reg}^\cdot$ from $M$.
For evaluation, the predicted and ground truth image scores are binarized using thresholds $\thresh{reg}^\cdot$
respectively $\thresh{GTreg}^\cdot=0.5$.
%
The average pooling kernel size $\constant{ksize}$
of both $\Pred{GT}_\text{reg}^\text{peaks}$ and $M_\text{reg}^\text{peaks}$
is set to 33\,pixels\footnote{
    Preliminary experiments showed: For $\constant{ksize}=2^i+1, i\leq 5$, differences between $\constant{ksize}$ for $\bPred{GT}_\text{reg}$ and $M_\text{reg}$ only had little influence on the results.
},
which approximately coincides with the typical height of a head in the test data
(cf.~size statistics in \cite{schwalbe_verification_2021})
at acceptable memory consumption.
This setting means, if half of the pixels within any $33\times33$\,pixels window
are false negatives, the image is marked ground truth faulty.
In our setting, \SI[round-precision=1]{27.7}{\percent} of the
test images are marked faulty.
For the Bool baseline, $\thresh{Bool}$ was set to $\thresh{px}$.

\subsubsection{Comparing Monitor Formulations}\label{sec:exp.fuzzylogic.formulations}
Compared were F1 score, precision, recall and true negative rate of different pixel- (\cref{eq:pixelmonitor}) and simple image-level (\cref{eq:monitoregionsimple}) monitor
formulations at $\thresh{px}=\thresh{reg}^\text{simple}=0.5$.
Aspects of variation were:%
\begin{itemize*}[label={}, itemjoin={;}, itemjoin*={; and}]
\item fuzzy logics from \cref{tab:fuzzylogics}
\item standard R-implication versus strong implication
\item adding calibration
\item adding denoising of masks, \idest setting concept mask values $<0.005$ to 0
\item bilinear upscaling versus maxpool downscaling of concept masks
\item the trivial $\Pred{CloseBy}_{\sigma,r}$ from \cref{eq:closeby}
    versus (only for upscaling) the non-trivial one with
    $\sigma\approx 2.77$ (truth value of 0.8 at distance of 4px) at
    $r=12$px (window of $25\times25$px, cutting at truth values below 0.1)
\end{itemize*}.
Preliminary experiments showed that R-implication produces many false positives at small input truth values, even with denoising.
Slight performance improvements were achieved by:%
\begin{itemize*}[label={}, itemjoin={;}, itemjoin*={; and}]
\item fuzziness
\item calibration for S-implication
\item the memory-intense $\Pred{CloseBy}_{\sigma\neq0}$ on pixel-level
\item downscaling for \L{} and P logic, at the cost of truth mask resolution
\end{itemize*}.
Therefore, we used the initially described setup for all later experiments. 


\subsubsection{Self-supervised Error Monitoring Results}\label{sec:exp.fuzzylogic.monitor}

Compared were Bool, \L{}, and P logic (cf.~\cref{tab:fuzzylogics}),
each with and without calibration (cal),
both for pixel-level (\cref{eq:pixelmonitor}, \cref{tab:perpixelauc}) and image-level (\cref{eq:regionmonitoradvancedours}, \cref{tab:auc}) monitoring.
Results show that
\begin{enumerate*}[label=(\arabic*), itemjoin={;\ }, itemjoin*={; and\ }, font=\itshape]
\item good error identification performance can be achieved,
uncovering a substantial amount of detector errors
(cf.~\ref{contrib:eval.performance})
\item fuzzy evaluations slightly but consistently outperform the non-fuzzy ones
and substantially outperform them for precision-biased metrics like \ol{F0.1} score
(cf.~\ref{contrib:eval.calibrationandfuzziness}),
\item calibrated slightly outperform non-calibrated versions
(cf.~\ref{contrib:eval.calibrationandfuzziness})
\end{enumerate*}.
The image-level results (\cref{tab:auc}) further reveal that
\begin{enumerate*}[resume, label=(\arabic*), itemjoin={;\ }, itemjoin*={; and\ }, font=\itshape]
\item acceptable precision-recall balances can be achieved for EffDet
(\forexample recall of $\geq 0.98$ at precision $\geq 0.60$,
or precision of $\geq 0.95$ at recall $\geq 0.1$)
\end{enumerate*}.
This can be seen on the precision-recall curves,
as well as from results for F0.1 score (favoring precision)
and F10 score (favoring recall) in \cref{tab:auc}.
Another finding was that
\begin{enumerate*}[resume, label=(\arabic*), itemjoin={;\ }, itemjoin*={; and\ }, font=\itshape]
\item the optimal $\thresh{reg}^\text{peaks}$ alarm threshold
consistently diverged from the default value 0.5
\end{enumerate*}. 
This suggests that tuning of the threshold on a validation set, similar to $\thresh{sIoU}$,
may boost performance values.
Without tuning, fuzziness brings a considerable benefit (cf.~$\text{F1}_{0.5}$ in \cref{tab:auc}).
This is especially of interest for cases when threshold tuning is not viable,
e.g., because of tuning data availability or bad generalization of the tuned threshold.

We also considered monitoring of the prediction-level formula for \enquote{A person should have a body part}
which should reveal false positives of the object detector.
With $\isPred{person,i}$ refering to the $i$th prediction for $P$:
\begin{align}
    F(P) = \left(\exists p\in P\colon \isPred{person,i}(p)\right) \rightarrow \big(\exists p\in P\colon \isPred{person,i}(p) \wedge
    \Pred{IsBodyPart}(p)
    \big)\;.
    \label{eq:fpformula}
\end{align}
A predicted box was defined as false positive (i.e., monitor GT positive)
if its person class score was greater 0.5 and it was covered by less than 20\,\% by the union of GT boxes,
producing positive rates of 0.03\,\% (EffDet) and 1.60\,\% (MR).
Evaluation was conducted using the P logic cal setting from \cref{tab:auc}.
Here, also, fuzzy approaches outperformed the Boolean one without threshold tuning (\cref{tab:resultsfprule}).
While performance was low for very small predicted boxes due to the static $\constant{ksize}$ setting and strong class imbalance, it achieved formidable F-scores on larger sized boxes,
especially for recall balanced settings which are relevant, \forexample, for sample selection for manual analysis.

\subsubsection{Comparing Logical Consistency of Models}\label{sec:exp.fuzzylogic.globalconsistency}
The better performing and calibrated (cf.~\cref{tab:modelresults}), and ca.\ three times larger MR model
also gives rise to better concept models (\cref{tab:caresults.perconcept}),
and achieves higher global logical consistency scores (\cref{tab:globalscore}).
with respect to the the rule from \cref{eq:fnformula}.
This aligns with the findings in \cref{tab:auc,tab:perpixelauc},
where fewer---but still a considerable amount---of the MR false negatives could be recovered using a logical consistency monitor.
On the other hand, most errors of the smaller EffDet were recovered using our simple example rule,
attesting the model some simple logical gaps.
This promises high potential for improvement of EffDet performance on person detection
by fine-tuning towards better logical consistency, or by post-processing based on the monitor outputs.


\subsubsection{Corner Case Analysis}\label{sec:exp.fuzzylogic.cornercases}
Finally, we manually inspected the samples with smallest mean pixel-wise logical consistency score for rule \cref{eq:fnformula}\footnote{
    Precisely, images $P_\text{img}$ with highest $M_\text{reg}^\text{simple}(P)$ value from \cref{eq:monitoregionsimple} for the region $P=\{ p\in P_\text{img}\mid M(p)\geq\num{e-3}\} \subset P_\text{img}$.
}.
Here, we exclude pixels $p$ regarded trivial (\forexample background), \idest such with low alarm score $M(p)\leq\num{e-3}$.
See \cref{fig:cornercases} for example outputs.
The analysis revealed for Mask~R-CNN:
\begin{enumerate*}[label=(\arabic*), itemjoin={,\ }, itemjoin*={, and\ }, font=\itshape]
\item persons segmented by (self-)occlusions may get too small bounding boxes
\item person features seem to be confused with that of animals and puppets
\end{enumerate*}.
And
\begin{enumerate*}[resume, label=(\arabic*), itemjoin={,\ }, itemjoin*={, and\ }, font=\itshape]
\item EffDet produces false negatives if the face is only slightly occluded,
\forexample by perspective, objects or the image boundary.
\end{enumerate*}
The found symbolic error modes directly allow to define data augmentation strategies,
\forexample adding persons segmented by occlusion.
To automate corner case selection from new samples, a threshold $\thresh{reg}^\cdot$ for the logical inconsistency score could be fine-tuned on a validation set.
Altogether, the fuzzy rule evaluation helps in finding semantic error modes.

\subsection{Limitations}\label{sec:exp.limitations}
Result quality depends on that of the framework ingredients:
the concept models and the rules.
Bad convergence,
a bias in the concept samples, or
simply insufficient CNN representations
may lead to erroneous concept masks.
And the prior knowledge rules are naturally susceptible to human bias (\forexample anatomical assumptions),
potentially leading to unfair safety guarantees.
Besides that, inconsistencies with those pre-defined rules are not guaranteed to accurately correspond to errors.
Our approach potentially considerably reduces the amount of test data needed to uncover issues
and it enables self-supervision,
but, thus, naturally inherits all limitations of data based verification methods, like test data representativity.
And lastly, parts of our simple initial modeling approach may be improved.
This concerns the simplistic CNN output transformation, GT definition,
and the neighborhood condition from \cref{eq:neighborcond1} which currently does not take into account object sizes, so still cannot prevent all occurrences of \enquote{over-shooting}.

\section{Conclusion and Outlook}

This work presents a simple, yet flexible, post-hoc, self-supervised method to verify and monitor outputs of trained CNNs regarding compliance with symbolic domain knowledge rules.
It is the first that requires no architectural constraints or changes to already trained CNNs,
and permits later extension of the rule base.
The method comes with little computational overhead, and
allows to leverage fuzziness and calibration for slight performance benefits.
We showed how to use it to identify a considerable proportion of detection errors.
Furthermore, the used logical consistency scores can help in finding error modes,
and to directly compare CNNs.
For example, in comparison to EfficientDet~D1, Mask~R-CNN shows considerably better representations and conformity with a simple occlusion robustness rule for person detection.

We are looking forward to see results on further models, tasks, and example rules,
including investigation of how to model temporal aspects, and how to advance our initial modeling suggestions.
Also, it seems promising to study how much a trained model could be improved by
self-supervised fine-tuning via the logical consistency scores.

\section*{Acknowledgments}
The research leading to these results is partly funded by the German
Federal Ministry for Economic Affairs and Energy within the project
\enquote{KI Absicherung – Safe AI for Automated Driving}. The authors
would like to thank the consortium for the successful cooperation.

%
%
\bibliographystyle{splncs04}
\bibliography{literature}

\clearpage
\appendix
\section*{Appendix Overview}

This supplemental material collects additional results and details for the experiments described in \themainpaper{}.
Specifically, it gives further evidence for the following claims made in the main paper:
\begin{itemize}
    \item \cref{sec:appendix.closeby}:
    \emph{The main formula for false negative detection used in \themainpaper{} can be implemented as parallelizable windowed operation} under some modeling constraints.
    \item \cref{sec:appendix.formulations}:
    This section gives detailed results of the preliminary experiments on the comparison of monitor formulations referenced in \themainpaper{}. These show that
    \emph{the simple rule formulation used in \themainpaper{} is a valid choice} for the later experiments in \themainpaper{}.
    \item \cref{sec:appendix.f1bythresh}:
    \emph{Fuzziness improves the error-detection performance of a logical consistency monitor when the hyperparameter for the binarization threshold cannot be tuned} on a validation set.
    \item \cref{sec:appendix.ksize}:
    For the peak-concentrated image-level monitor formulation,
    \emph{the kernel size hyperparameter of the monitor can be chosen independently of that of the ground truth.}
\end{itemize}

\cref{sec:appendix.hyperparams} gives details on the chosen hyperparameter values for the experiments to ensure reducibility.

\begin{table*}
    \centering
    \footnotesize
    \caption{Overview of hyperparameters and their default values used in the formula calculation (here and in \themainpaper{}).}
    \begin{tabular}{@{} >{\ttfamily} l l l@{}}
        \toprule
        \multicolumn{1}{@{}l}{Name} & Default & Function
        \\\midrule[\heavyrulewidth]
        $\thresh{sIoU}$  & 0.5   & Threshold used to determine the set intersection over union performance
        \\$\thresh{denoise}$ & 0.005 & Denoise low values of $\Pred{IsBodyPart}(\bullet)$ masks
        \\$\thresh{ped}$& 0.5   & Binarizing threshold for $\isPred{person}(\bullet)$ masks
        \\$\thresh{px}$ & 0.5   & Binarizing threshold for pixel-level monitor output masks
        \\$\thresh{reg}^\cdot$ & 0.5   & Binarizing threshold for image-level monitor outputs
        \\$\thresh{sIoU}$ & 0.5 & Binarizing threshold for pixel-level monitor outputs for measuring sIoU
        \\$\thresh{Bool}$    & 0.5   & Binarizing threshold for all mask in non-fuzzy logic
        \\ksize    & 33    & Width of the quadratic kernel for defining a (binary) pixel neighborhood
        \\\bottomrule
    \end{tabular}
    \label{tab:constants}
\end{table*}

\section{Implementation Notes: Hyperparameter Choices}\label{sec:appendix.hyperparams}
In the following, hyperparameter choices for our experiments are detailed.
An overview on further hyperparameters introduced in this supplemental material and \themainpaper{} can be found in \cref{tab:constants}.

Settings applicable to all experiments are:
\begin{itemize}
\item An image size of $400\times400$ pixels was used for Mask~R-CNN, $640\times640$ pixels for EfficientDet~D1.
\item Images were first zero-padded to square size, then resized to the desired image size (keeping the aspect ratio).
\item For metrics working on binary pixel values, output masks (after transformation of bounding boxes to masks) are thresholded at a value of 0.5.
\end{itemize}

\paragraph{Concept Data}
The body part segmentation masks were generated from keypoint annotations as proposed in \cite{schwalbe_verification_2021}.
Point concepts (\Concept{eye}, \Concept{wrist}, \Concept{ankle}) are added as filled circle of white pixels. Limb concepts (\Concept{arm}, \Concept{leg}) get filled circles at edges and joints, and lines connecting them where skeletal connections exist.
The concepts consist of the following keypoints:
\begin{itemize}
    \item \Concept{arm}: left/right shoulder, ellbow, wrist
    \item \Concept{leg}: left/right hip, knee, ankle
    \item \Concept{eye}, \Concept{wrist}, \Concept{ankle}: corresponding left/right keypoints
\end{itemize}
Circle diameter and line width were set to \SI{5}{\percent} of the estimated person body height in pixels (using estimation algorithm from \cite{schwalbe_verification_2021}).
The body height defaults to the bounding box height if it cannot be estimated from keypoint links of the annotation.
Keypoints annotated as occluded are treated as not present.

\paragraph{Calibration and Formula Evaluations}
The measurement of calibration and performance metrics on the main CNN, as well as the evaluations of the fuzzy logic formulas, used the following settings:
\begin{itemize}
    \item Evaluation batch size: 64;
    \item Data split: All images from the MS~COCO dataset~\cite{lin_microsoft_2014} val2017 split are used for evaluation.
\end{itemize}

\paragraph{Concept Model Training}
The training and evaluation of the concept models used the following settings in accordance with \cite{schwalbe_verification_2021}:
\begin{itemize}
    \item Metric and loss calculation:
    Concept model outputs (after convolution) are first bilinearly upscaled to match the ground truth resolution, then normalized.
    For BCE-losses, normalization is skipped and the loss is calculated in logit space.
    \item Loss for second stage training of the Laplace approximation parameters \cite{mackay_evidence_1992}: binary cross-entropy (in logit space);
    \item Data split: The training and validation data sets are taken from the MS~COCO dataset~\cite{lin_microsoft_2014} train2017 split, the test data from the val2017 split.
    For each concept, only those samples are included that contain any positive concept mask pixel.
    The train2017 samples for a concept are randomly split into training and validation set, at a ratio of 4:1.
    \item Optimizer and learning rate: Adam~\cite{kingma_adam_2015} from PyTorch version 1.9.0 implementation, with a learning rate of 0.001, and default beta values of 0.9 and 0.999; no weight decay;
    \item Batch sizes: 8 for training, 64 for validation, and 6 for the second stage training of the Laplace approximation parameters \cite{mackay_evidence_1992};
    \item Early stopping: Concept models are trained for at most 7 epochs. Training is stopped early after an epoch if the validation loss decreases less than 0.001 for each in three successive epochs.
\end{itemize}

\section{Implementation Notes: Parallelization of Example Formula}\label{sec:appendix.closeby} 
Consider the mask operation $\Pred{CloseToA}_\text{\Concept{b}}$ that accepts a mask $Q$ and returns a mask $P$ with each pixel $p$ holding the truth value for \emph{\enquote{$p$ is close to a pixel $q$ in $Q$ at which concept \Concept{b} is present}}.
This is used in the example formula from \themainpaper{} as will become clear from \cref{def:closetoa}.
This section shows that under some conditions this
\textbf{$\Pred{CloseToA}_\text{\Concept{b}}$ can be implemented as a windowed operation} on the mask $Q$, similar to a convolution or pooling operation. Furthermore, an upper bound is given for the minimal window size (cf.~\cref{prop:closetoawindowable}).
The \textbf{windowing allows parallelization, and thus a potentially massive speedup}.
Note, however, that this may be quite memory intensive, as it is not guaranteed that the operation can be represented as sparse matrix operations.

Before formally defining $\Pred{CloseToA}_\text{\Concept{b}}$ in \cref{def:closetoa}, some notation for pixel indices is introduced that allows to interpret them as coordinates (with standard distance measures like $L_1$).
\begin{Def}
    Let $h,h',w,w'$ be natural numbers (mask heights and widths), and let $Q\in[0,1]^{h'\times w'}, P\in[0,1]^{h\times w}$ be 2D masks.
    A mask coordinate system here denotes an injective mapping of dimensions $(i_h,i_w)$ in a mask $P$ to pixel positions in 2D space $\mathds{R}^2$.
    By default map the dimension $(i_h,i_w)$ to the point $(i_h+0.5, i_w+0.5)$, \idest a pixel has a width of $1\times1$ and pixel coordinates refer to pixel centers.
    Let $p\in\mathds{R}^2$ be a pixel position.
    \begin{enumerate}[label=(\alph*)]
    \item \emph{Pixel positions in a mask:}
        Denote by $p\in P$ that $p$ is a pixel in the mask $P$.
    \item \emph{Pixel values of a mask:}
        Given that $p\in P$, denote by $P(p)$ the value at the pixel position $p$ in the mask $P$.
    \item \emph{Sub-masks of a mask:}
        A mask $W\in[0,1]^{h\times w}$ is a window or sub-mask in $P\in[0,1]^{h'\times w'}$, denoted $W\subset P$,
        if $h\leq h', w\leq w'$, and there is a mask coordinate system for $W$ and $P$ such that for all pixel positions $p\in W$ holds $p\in P$ and $W(p)=P(p)$.
    \end{enumerate}
\end{Def}
\begin{Def}\label{def:closetoa}
     Let $h,h',w,w'$ be integers and $Q\in[0,1]^{h'\times w'}$ be a mask.
     Assume a logic is given with predicates $\isPred{b}$ and $\Pred{CloseBy}$ of arity one respectively two, which both operate on pixel positions.
     We define the following mask operation:
    \begin{align}
        \Pred{CloseToA}_\text{\Concept{b}}&\colon [0,1]^{h'\times w'}\to[0,1]^{h\times w} \\
        P(p)&\coloneqq \left(
            \exists q\in Q\colon \isPred{b}(q) \wedge \Pred{CloseBy}(p, q)
        \right)
    \end{align}
    for $p\in P\coloneqq\Pred{CloseToA}_\text{\Concept{b}}(Q)$.
\end{Def}
This is used in the example formula from \themainpaper{}.
(Note that $P$ and $Q$ need not have the same resolution.)

The following proposition states that, under some conditions,
$\Pred{CloseToA}_\text{\Concept{b}}$ can be implemented as a windowed operation.
And, if the $\Pred{CloseBy}$ predicate becomes zero at some $L_1$ distance $r$, the window size of the windowed operation may be chosen $2r+1$ (or smaller). This means, $2r+1$ is the maximum window size necessary to ensure an exact implementation.

\begin{Prop}\label{prop:closetoawindowable}
    Consider a (fuzzy) logic, and $\isPred{b}$, $\Pred{CloseBy}$, and the masks $Q$ and $P=\Pred{CloseToA}_\text{\Concept{b}}(Q)$ as in \cref{def:closetoa}.
    Assume the following:
    \begin{enumerate}[label=(\alph*), ref=(\alph*)]
        \item\label{prop:closetoawindowable.exists}
            $\exists$ is either defined as an arithmetic $\mean$ \cite{donadello_logic_2017}, or defined using the logical \emph{OR},
            \idest for a domain $X$ and a formula $f$ holds
            $\exists x\in X\colon f(x) \coloneqq \bigvee_{x\in X}f(x)$.
        \item\label{prop:closetoawindowable.closebyzero}
        There is a square integer window size $2r+1$ such that $\Pred{CloseBy}(p,q)=0$ if $\|p-q\|_1>r$ for any pixel positions $p,q$.
        This is \forexample the case for the suggested maxpooling or Gaussian $\Pred{CloseBy}$ from \themainpaper{}.
    \end{enumerate}
    Then, $P(p)$ for a pixel position $p\in P$ at most depends on the pixel values in the window $W_p\subset Q$ of size $(2r+1)\times(2r+1)$ centered at $p$, \idest
    \begin{align}
        \Pred{CloseToA}_\text{\Concept{b}}(Q)(p)
        = \left(
            \exists q\in W_p\colon \isPred{b}(q) \wedge \Pred{CloseBy}(p, q)
        \right)
    \end{align}
    \begin{proof}
    Let $f(p,q)\coloneqq \isPred{b}(q) \wedge \Pred{CloseBy}(p, q)$.
    Note that
    \begin{enumerate}[nosep,label=(\roman*), ref=(\roman*)]
    \item\label{prop:closetoawindowable.proof.i} $f(p,q)=0$ if $\Pred{CloseBy}(p, q)=0$, and
    \item\label{prop:closetoawindowable.proof.ii} $q\in W_p$ if $f(p,q)\neq 0$,
        because for $q\in Q\setminus W_p$ holds:
        \begin{itemize}[nosep]
        \item $\|p-q\|_1>r$ by definition of the window $W_p$, thus
        \item $\Pred{CloseBy}(p, q)=0$ by \ref{prop:closetoawindowable.closebyzero}, and so
        \item $f(p, q)=0$ by (*).
        \end{itemize}
    \end{enumerate}
    For $\exists$ defined via logical \emph{OR} according to
    assumption \ref{prop:closetoawindowable.exists}, the formula becomes
    \begin{align}
        P(p)
        &= \exists q\in Q\colon f(p, q)
        \overset{\text{\ref{prop:closetoawindowable.exists}}}{=} \bigvee_{q\in Q} f(p, q)
        \\&\overset{\text{\ref{prop:closetoawindowable.proof.i}}}{=}
        0 \vee \bigvee_{\mathclap{\substack{q\in Q\\f(p, q)\neq 0}}} f(p,q)
        \overset{\text{\ref{prop:closetoawindowable.proof.ii}}}{=}
        0 \vee \bigvee_{\mathclap{\substack{q\in W_p\\f(p, q)\neq 0}}} f(p,q)
        \\&\overset{\text{\ref{prop:closetoawindowable.exists}}}{=}
        \exists q\in W_p\colon f(p, q)
    \end{align}
    For $\exists$ defined via arithmetic mean, note that
    \begin{align}
        \exists q\in Q\colon f(p, q) = \frac{1}{\#Q} \sum_{q\in Q} f(p, q)
        \;.
    \end{align}
    Replacing all occurrences of $\bigvee_{q\in Q}$ with $\frac{1}{\#Q} \sum_{q\in Q}$ in the above calculation finalizes the proof.
    \end{proof}
\end{Prop}
\begin{Remark}
    Note that the notation, \cref{def:closetoa}, and \cref{prop:closetoawindowable} all can easily be generalized to masks $Q\in[0,1]^{n_1\times\dots\times n_d}$ of arbitrary dimension $d\geq 1$, with pixel positions in $\mathds{R}^d$.
    So, the rule formulation and efficient implementation can be applied not only to 2D masks, but also \forexample 3D masks consisting of several channels or frames.
\end{Remark}

\begin{Remark}
    The following special cases further simplify the implementation:
    \begin{enumerate}[label=(\alph*)]
        \item \emph{Trivial case:}
            In case $r=1$,
            as in the experiments in \themainpaper{}, $\Pred{CloseToA}_\text{\Concept{b}}(Q)(p)=\isPred{b}(p)\wedge \Pred{CloseBy}(p,p)$.
        \item \emph{Maxpooling case:}
            For $\exists=\max$ and binary
            \begin{align}
                \Pred{CloseBy}(p, q)\coloneqq(\|p-q\|_1\leq r)
            \end{align}
            ($\Pred{CloseBy}(p,q)$ is one if $q$ lies in a square window around $p$, else 0)
            $\Pred{CloseToA}_\text{\Concept{b}}$ becomes a simple maxpooling operation on $Q$ with kernel size $2r+1$.
            The stride of the pooling operation is such that the size $h'\times w'$ of $Q$ is reduced to the size $h\times w$ of P. 
            This is used for the downscaling operation which was suggested in \themainpaper{} and is further investigated in \cref{sec:appendix.formulations}.
        \item \emph{Convolutional case:}
            For $\exists=\mean$ \cite{donadello_logic_2017} and Product logic,
            $\Pred{CloseToA}_\text{\Concept{b}}$ becomes a simple linear convolution followed by applying a pixel-wise factor $\frac{1}{(h'\cdot w')^2}$.
            Kernel weights for the kernel window with center $p$ are defined by $\Pred{CloseBy}(p, \cdot)$.
    \end{enumerate}
\end{Remark}

\section{Comparison of Monitor Formulations}\label{sec:appendix.formulations}
This section collects further considerations and the results for directly comparing different monitor formulations.
The following variations are considered (cf.~\cref{tab:formulavariations}):
\begin{description}[font=\normalfont\itshape]
    \item[Fuzzy logics] from \themainpaper{}: \L{}ukasiewicz, Product, and Goedel fuzzy logic; and non-fuzzy predicate logic on concept masks binarized at $\thresh{Bool}$.
    \item[Implication style:] Sstrong implication (S-implies) or residuated implication (R-implies).
    \item[Denoising:]
    With or without setting concept mask values lower than $\thresh{denoise}=0.005$ to 0.
    This can be modeled as pixel-wise operation $(\isPred{b}(p)\geq \thresh{denoise}) \wedge \isPred{b}(p)$ for pixels $p$ and concept $\Concept{b}$ with concept mask $\isPred{b}(\cdot)$.
    \item[Choice of $\Pred{CloseBy}$:]
    Considered are the Gaussian $\Pred{CloseBy}_{\sigma, r}$ from \themainpaper{}, different mask scaling options, and trivial $\Pred{CloseBy}_{\sigma=0,r=1}$.
    The following three settings are compared here:
    \begin{itemize}[nosep]
        \item bilinear upscaling with trivial $\Pred{CloseBy}$
        \item bilinear upscaling with non-trivial $\Pred{CloseBy}_{\sigma,r}$, $\sigma\approx2.77$ and $r=12$px
        (\emph{in visualizations shorted to $\Pred{CoveredBy}$})
        \item downscaling using maxpooling with trivial $\Pred{CloseBy}$
    \end{itemize}
    \item[Calibration:]
    With or without post-hoc Laplace calibration of concept model outputs.
\end{description}

\paragraph{On the Denoising}
The \emph{denoising} intends to reduce the instable cases of pixels for R-implies.
These are the cases where both values of $\Pred{IsBodyPart}$ and $\isPred{person}$ are very low, but $\Pred{IsBodyPart}$ is larger than $\isPred{person}$.
For Product and Goedel logic this causes $\Pred{IsBodyPart}(p)\rightarrow_R\isPred{person}(p)$ to jump from 1 to a smaller value (cf.~R-implies definitions from \themainpaper{}).

\paragraph{On the $\Pred{CloseBy}$ Choice}
Either approach for \emph{implementing $\Pred{CloseBy}$}---the fuzzy but memory-intense Gaussian $\Pred{CloseBy}_{\sigma, r}$, and the non-fuzzy maxpooling based one---has the effect
of enlarging the area of positive truth values.
This is useful if concept masks have different resolutions and might not match precisely at the boundaries. This leads to low truth values in boundary regions if a low-resolution mask and a higher-resolution mask are scaled and combined in a \emph{AND} or \emph{implies} operation.

\begin{figure*}
    \centering
    \includegraphics[width=\textwidth]{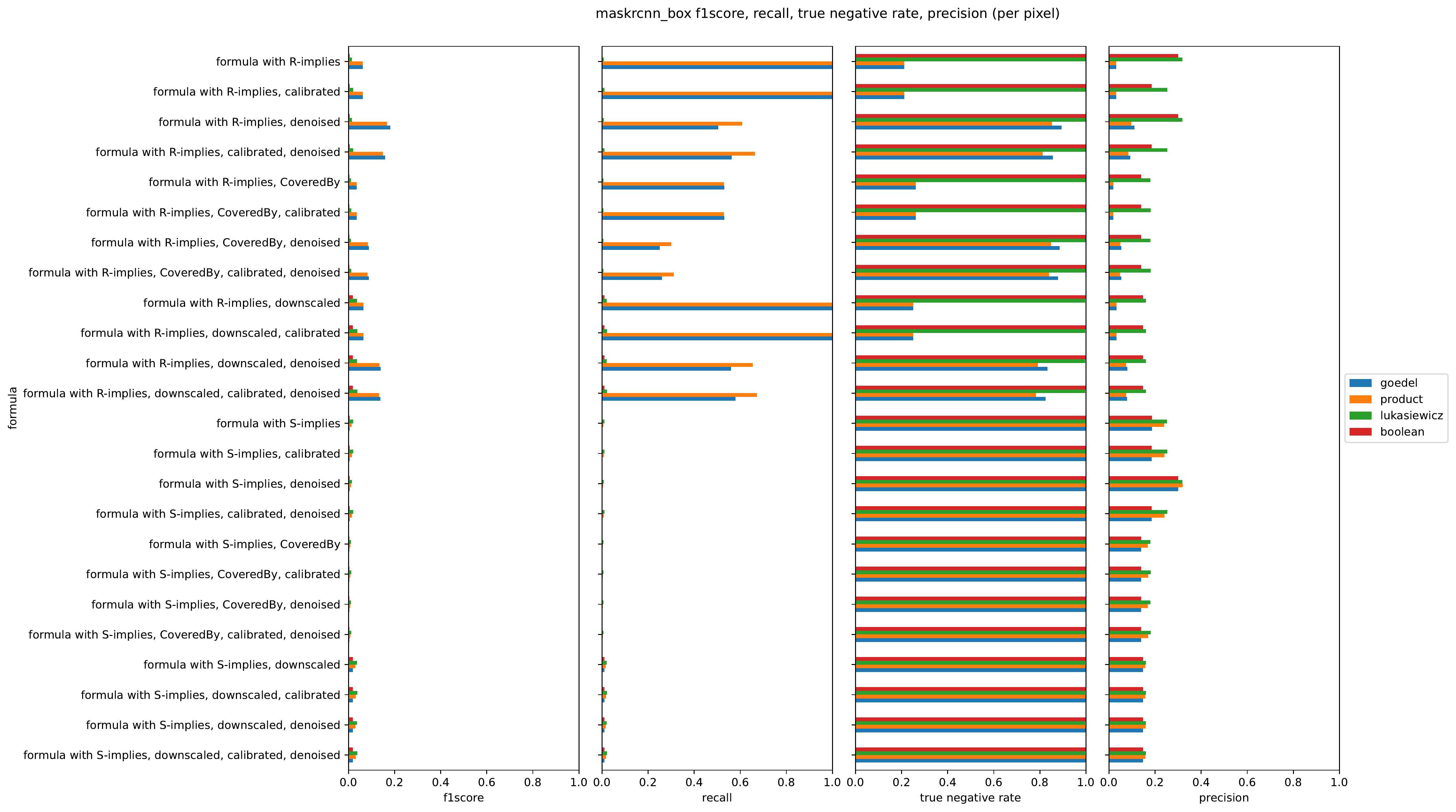}%
    \caption{
        \textbf{Mask R-CNN}, \textbf{pixel-level} monitor:
        Visual comparison of performance metrics for different monitor formulations (cf.~\cref{tab:formulavariations}).
        \emph{Left to right:} F1 score, recall, precision, and true negative rate.
    }
    \label{fig:perpixelmetricsmaskrcnn}
\end{figure*}

\begin{figure*}
    \centering
    \includegraphics[width=\textwidth]{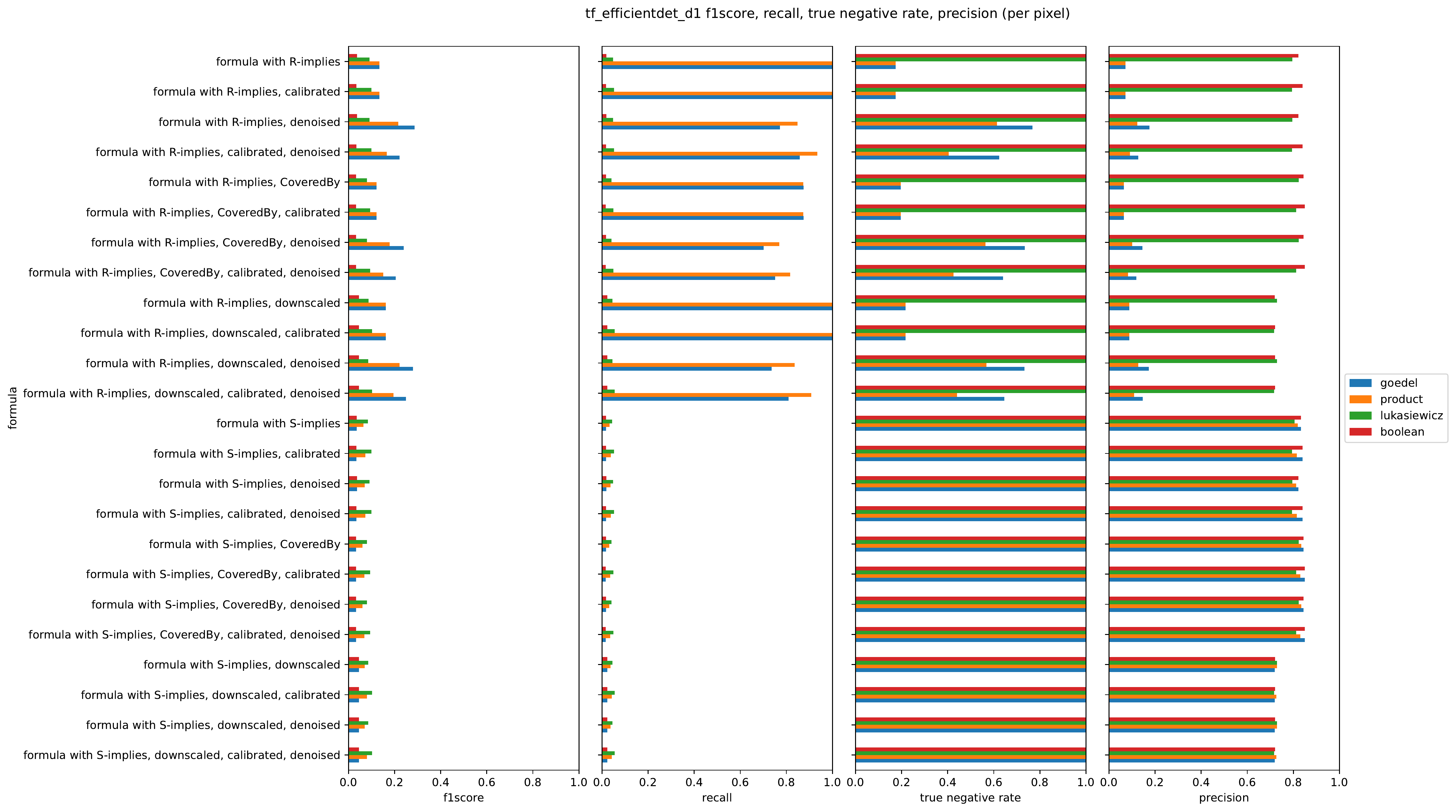}%
    \caption{
        \textbf{EfficientDet D1}, \textbf{pixel-level} monitor:
        Visual comparison of performance metrics for different monitor formulations (cf.~\cref{tab:formulavariations}).
        \emph{Left to right:} F1 score, recall, precision, and true negative rate.
    }
    \label{fig:perpixelmetricsefficientdet}
\end{figure*}

\begin{figure*}
    \centering
    \includegraphics[width=\textwidth]{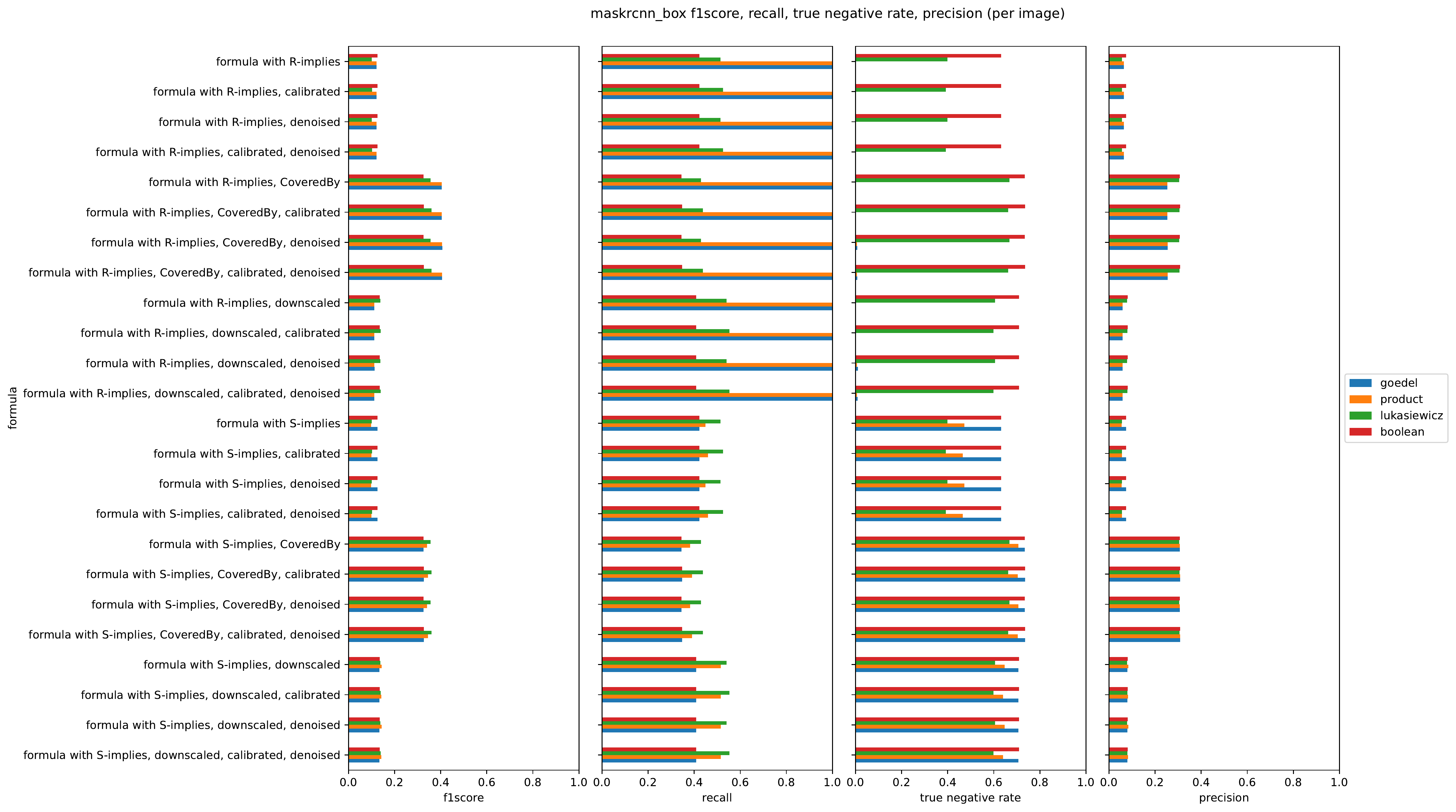}
    \caption{
        \textbf{Mask R-CNN}, \textbf{image-level} monitor:
        Visual comparison of performance metrics for different monitor formulations (cf.~\cref{tab:formulavariations}).
        \emph{Left to right:} F1 score, recall, precision, and true negative rate.
    }
    \label{fig:perimgotherscores/maskrcnn}
\end{figure*}

\begin{figure*}
    \centering
    \includegraphics[width=\textwidth]{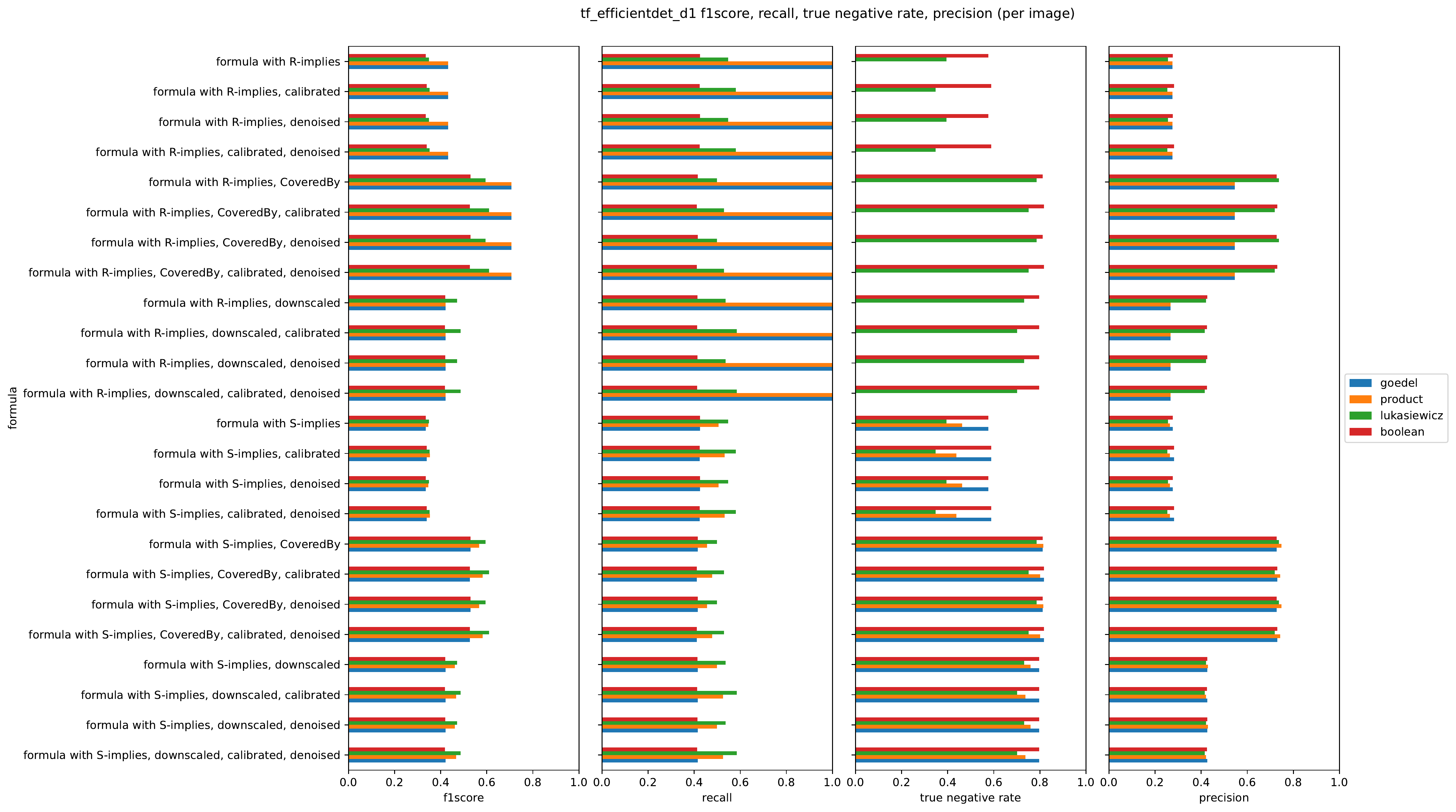}
    \caption{
        \textbf{EfficientDet D1}, \textbf{image-level} monitor:
        Visual comparison of performance metrics for different monitor formulations (cf.~\cref{tab:formulavariations}).
        \emph{Left to right:} F1 score, recall, precision, and true negative rate.
    }
    \label{fig:perimgotherscores/efficientdet}
\end{figure*}

\begin{figure*}
    \centering
    \includegraphics[width=0.95\linewidth]{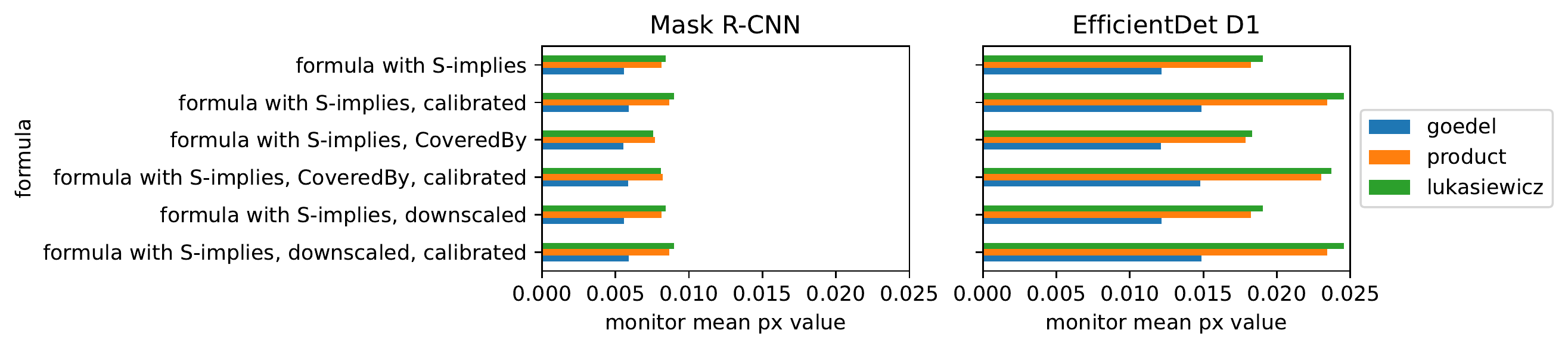}
    \caption{
        Comparison of values of $(1-\text{global score})$ for different monitor formulations.
        This calculates as the mean of the pixel-wise monitor values over the complete dataset.
    }
    \label{fig:globalscores}
\end{figure*}

\paragraph{Results}
Results for the following monitor formulations can be found below (both using binarization thresholds $\thresh{px}=0.5=\thresh{reg}^\text{simple}$):
\begin{itemize}
    \item
    Pixel-wise monitor $M$:
    \cref{fig:perpixelmetricsmaskrcnn} for Mask~R-CNN,
    \cref{fig:perpixelmetricsefficientdet} for EfficientDet~D1.
    \item 
    Image-level monitor $M_\text{reg}^\text{simple}$:
    \cref{fig:perimgotherscores/maskrcnn} for Mask~R-CNN,
    \cref{fig:perimgotherscores/efficientdet} for EfficientDet~D1.
\end{itemize}
Furthermore, global scores for different monitor formulations are compared in \cref{fig:globalscores}.
It must be noted that the pixel-level monitor is expected to have low recall: Only such detector false negative pixels should be highlighted that are inconsistent with the considered rule, which can only be those that are part of one of the considered body parts (no torso, bounding box corners, etc.).
The following can be seen from the visualizations:
\begin{enumerate}[label=(\arabic*), font=\itshape]
\item \emph{Fuzziness}:
    For S-implication, \textbf{non-fuzzy logic performs consistently slightly worse than fuzzy logic.}
\item \emph{Implication style} and \emph{denoising}:
    Despite denoising, \textbf{R-implication is still much too sensitive} and causes many false positives (cf.~low true negative rates, and suspiciously high recall on pixel-level).
\item \emph{$\Pred{CloseBy}$}:
    \textbf{Using a non-trivial $\Pred{CloseBy}$
    gives slight (downscaling) up to significant (non-trivial $\Pred{CloseBy}_{\sigma, r}$) performance benefits}, as can be seen \forexample from the F1 scores.
    However, $\Pred{CloseBy}_{\sigma, r}$ either cannot be parallelized or requires lots of memory for the windowed operation,
    and downscaling significantly reduces (localization) information.
    Thus, in \themainpaper{} only the trivial $\Pred{CloseBy}$ is used, and noise problems smoothed using the neighborhood condition for the image-level monitor (cf.~\cref{sec:appendix.ksize}).
\item \emph{Calibration}
    on pixel-level seems to have a \textbf{negative effect on precision and a positive one on recall, resulting in comparable F1 scores}.
    On image-level, calibration seems to have the least effect of all aspects of variation on the monitor performance.
    \textbf{For the global scores calibration has the visible effect of decreasing logical consistency for any rule formulation}.
\item The comparison of the lack of \emph{global logical consistency} in \cref{fig:globalscores} shows that \textbf{the choice of rule formulation has no influence on the trend of the global consistency}. Any choice of variation attests EfficientDet~D1 the clearly worse score.
\end{enumerate}

\begin{table*}
    \centering
    \caption{Overview on the different aspects of variation tested for the formulation of the formula from \themainpaper{} (cf.~\cref{sec:appendix.formulations})
    }
    \label{tab:formulavariations}
    \begin{tabularx}{\textwidth}{@{} l @{$\quad$} X @{}}
        \toprule
        Aspect & Description
        \\\midrule[\heavyrulewidth]
        Implication        & Whether S- or R-implication is used
        \\$\Pred{CloseBy}$, Scaling & Formulation of the $\Pred{CloseBy}$:
                            upscaling of small masks with either non- or trivial $\Pred{CloseBy}_{\sigma,r}$;
                            or $\max$-downscaling of large masks
        \\Calibration      & Whether Laplace calibration is applied to the concept models during inference
        \\Denoising        & Whether truth values in concept masks $\isPred{b}(\cdot)$ that are below $\thresh{denoise}$ are set to 0
        \\\bottomrule
    \end{tabularx}
\end{table*}

\section{Benefits of Fuzziness without Monitor Threshold Tuning}\label{sec:appendix.f1bythresh}
\begin{figure*}
    \centering
    \begin{subfigure}{0.48\linewidth}
        \includegraphics[width=\linewidth]{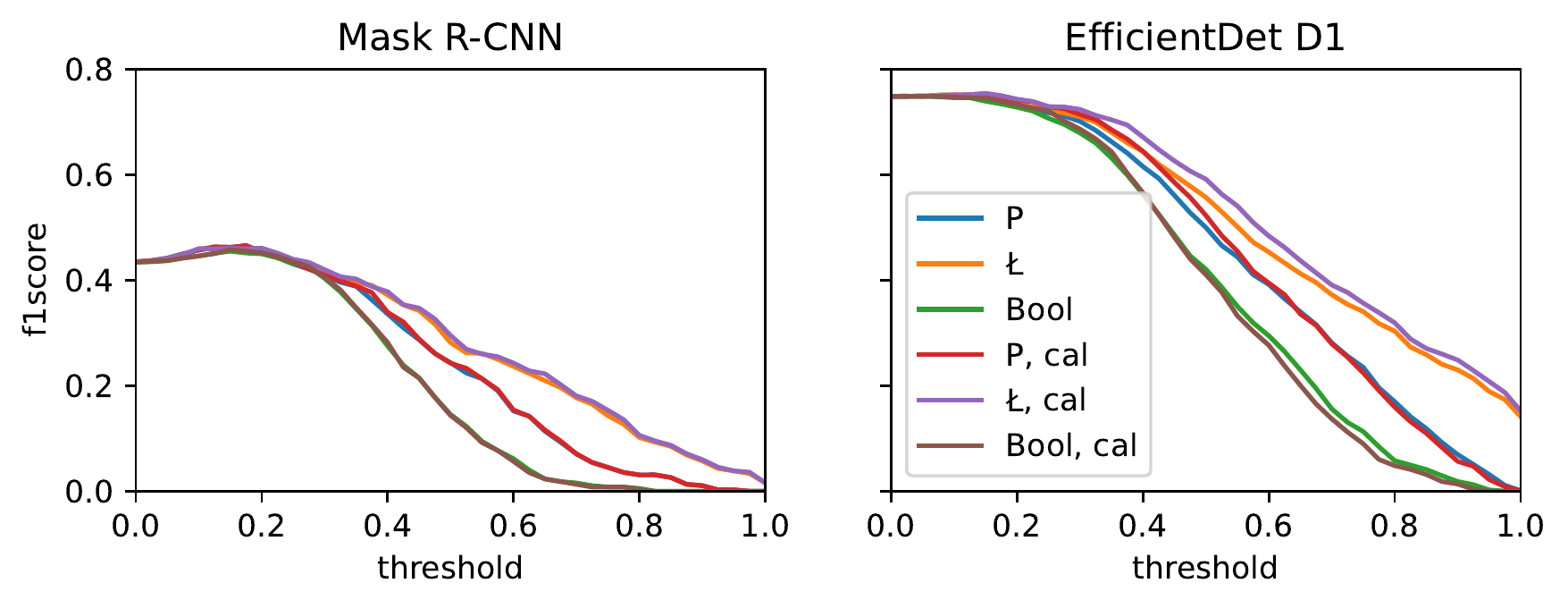}    
        \caption{Per-image F1 score by threshold}
        \label{fig:f1bythresh.perimg}
    \end{subfigure}%
    \hfill%
    \begin{subfigure}{0.48\linewidth}
        \includegraphics[width=\linewidth]{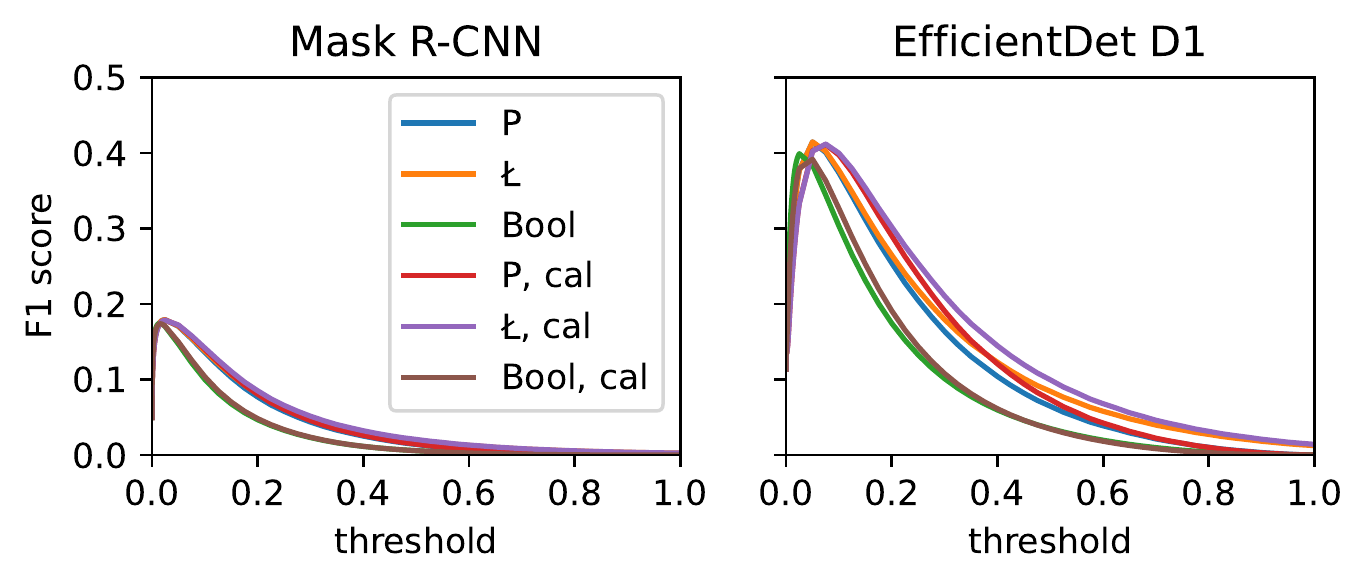}
        \caption{Per-pixel F1 score by threshold}
        \label{fig:f1bythresh.perpixel}
    \end{subfigure}
    \caption{
        F1 score by
        $\thresh{reg}^\text{peaks}$ (image-level, \cref{fig:f1bythresh.perimg}) and $\thresh{px}$ (pixel-level, \cref{fig:f1bythresh.perpixel})
        for the versions of the simple rule formulations compared in \themainpaper{}.
    }
    \label{fig:f1bythresh}
\end{figure*}

\cref{fig:f1bythresh} shows the F1 scores of the pixel- and the image-level monitors considered in \themainpaper{}, plotted by the binarization threshold $\thresh{px}$ respectively $\thresh{reg}^\text{peaks}$.
Concretely compared are the non-fuzzy baseline with \L{}ukasiewicz and Product t-norm fuzzy logic, with calibrated and non-calibrated concept models.
The following can be seen:
\begin{enumerate}[label=(\arabic*), font=\itshape]
    \item \textbf{When deviating from the optimal threshold, fuzzy formulations clearly outperform the non-fuzzy alternative.}
    \item The optimal thresholds, and corresponding optimal \ol{F1} scores on the test set nearly coincide for all formulations.
    \item In all cases, F1-by-threshold curve has a unique maximum,
    and the threshold for this optimal F1 score is far from the default of 0.5.
    This \textbf{highlights the potential performance increase by tuning the hyperparameter $\thresh{px}$ respectively $\thresh{reg}^{\cdot}$}.
    \item It can be seen a consistent slight performance benefit of calibration.
\end{enumerate}

\section{Choice of $\constant{ksize}$ for Image-level Monitor}\label{sec:appendix.ksize}
The peak-concentrated image-level monitor $M_\text{reg}^\text{peaks}$ used in experiments in \themainpaper{} has the
additional hyperparameter of the average pooling kernel size $\constant{ksize}$ for the image-level monitor neighborhood condition.
As both the monitor $M_\text{reg}^\text{peaks}$ and $\Pred{GT}_\text{reg}^\text{peaks}$
were defined using the neighborhood condition, each has a tunable $\constant{ksize}$ parameter, $\constant{ksize}_M$ and $\constant{ksize}_\text{GT}$.

\paragraph{Ground Truth Positive Rates}
$\constant{ksize}_\text{GT}$ influences the ground truth positive rate of images in the dataset, as smaller kernel window size values are more sensible to smaller (possibly noisy) peaks. \cref{tab:perimggtpositiverate} shows the positive rates for exponentially increasing $\constant{ksize}_\text{GT}$ and two $\thresh{GTreg}^\text{peaks}$ settings.
What positive rate is obtained for which $\constant{ksize}$ setting depends on the distribution of alarm area sizes in the test dataset.
For example, for $\thresh{GTreg}^\text{peaks}$ the positive rate approximately linearly increases with the kernel size.
Note that for $\constant{ksize}_\text{GT}$, this is identical with the ground truth for the simple monitor formulation $M_\text{reg}^\text{simple}$.

\begin{figure*}
    \centering
    \mbox{
    \includegraphics[width=.6\textwidth]{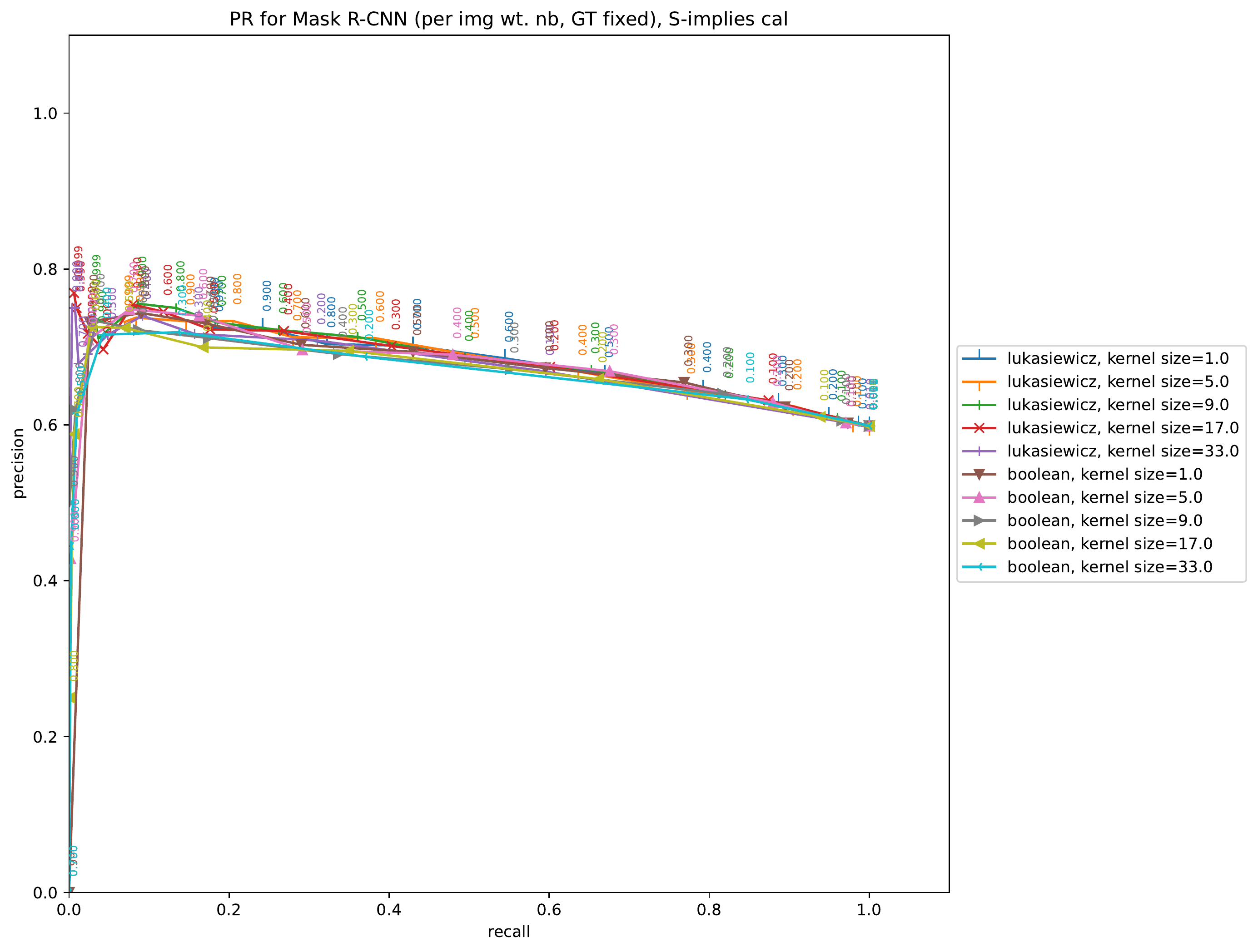}%
    \hspace*{-0.18\textwidth}%
    \includegraphics[width=.6\textwidth]{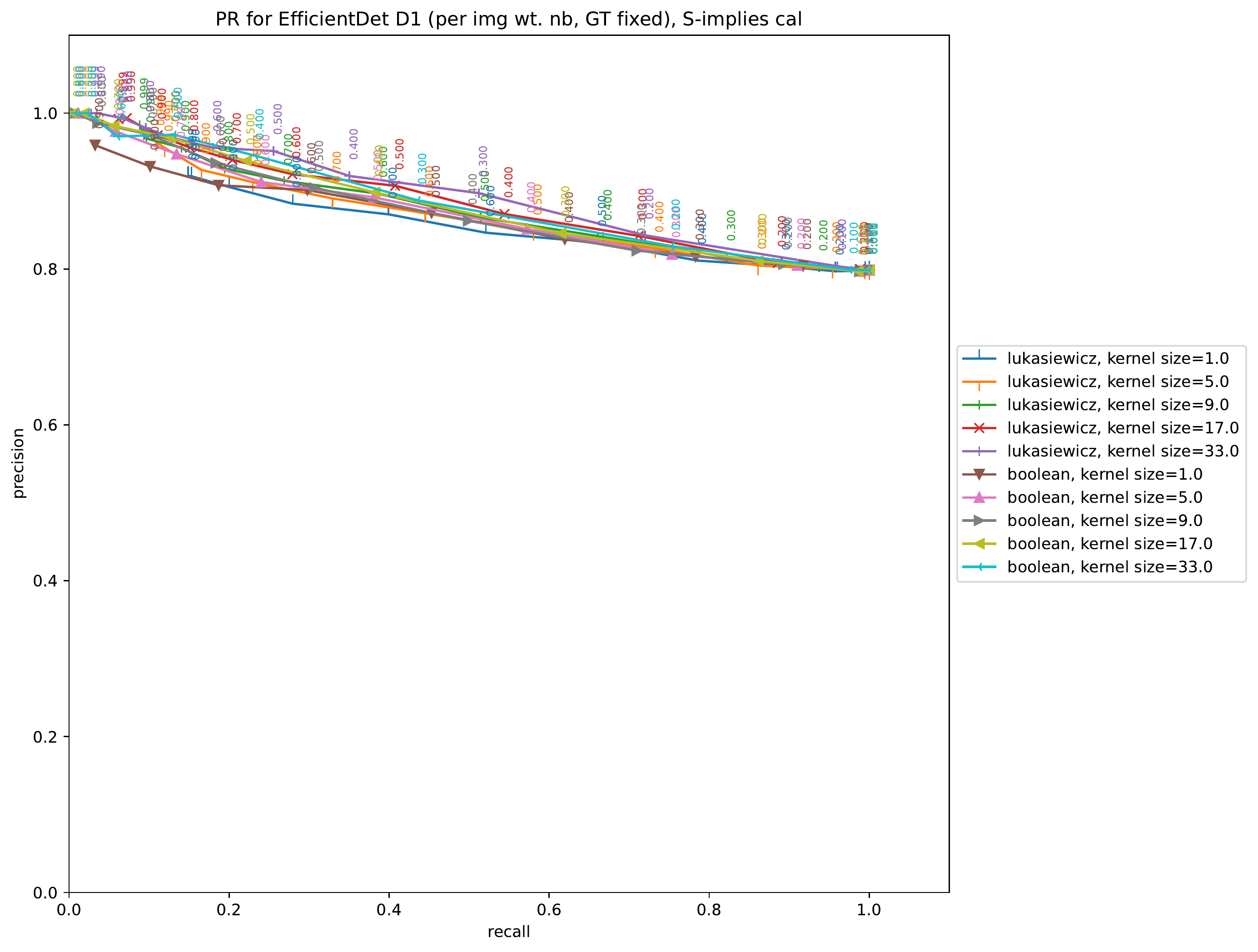}%
    }\\%
    \mbox{
    \includegraphics[width=.6\textwidth]{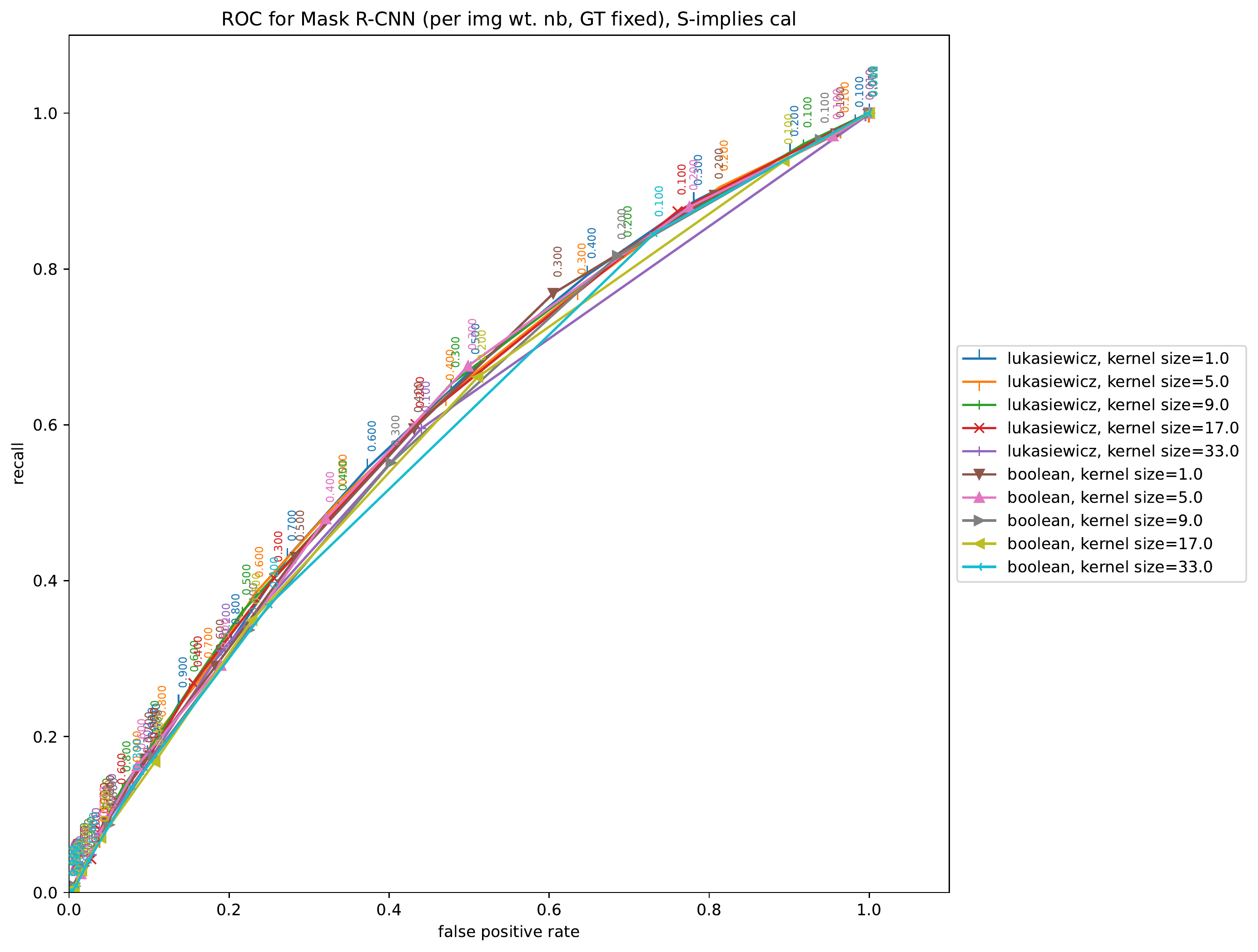}%
    \hspace*{-0.18\textwidth}%
    \includegraphics[width=.6\textwidth]{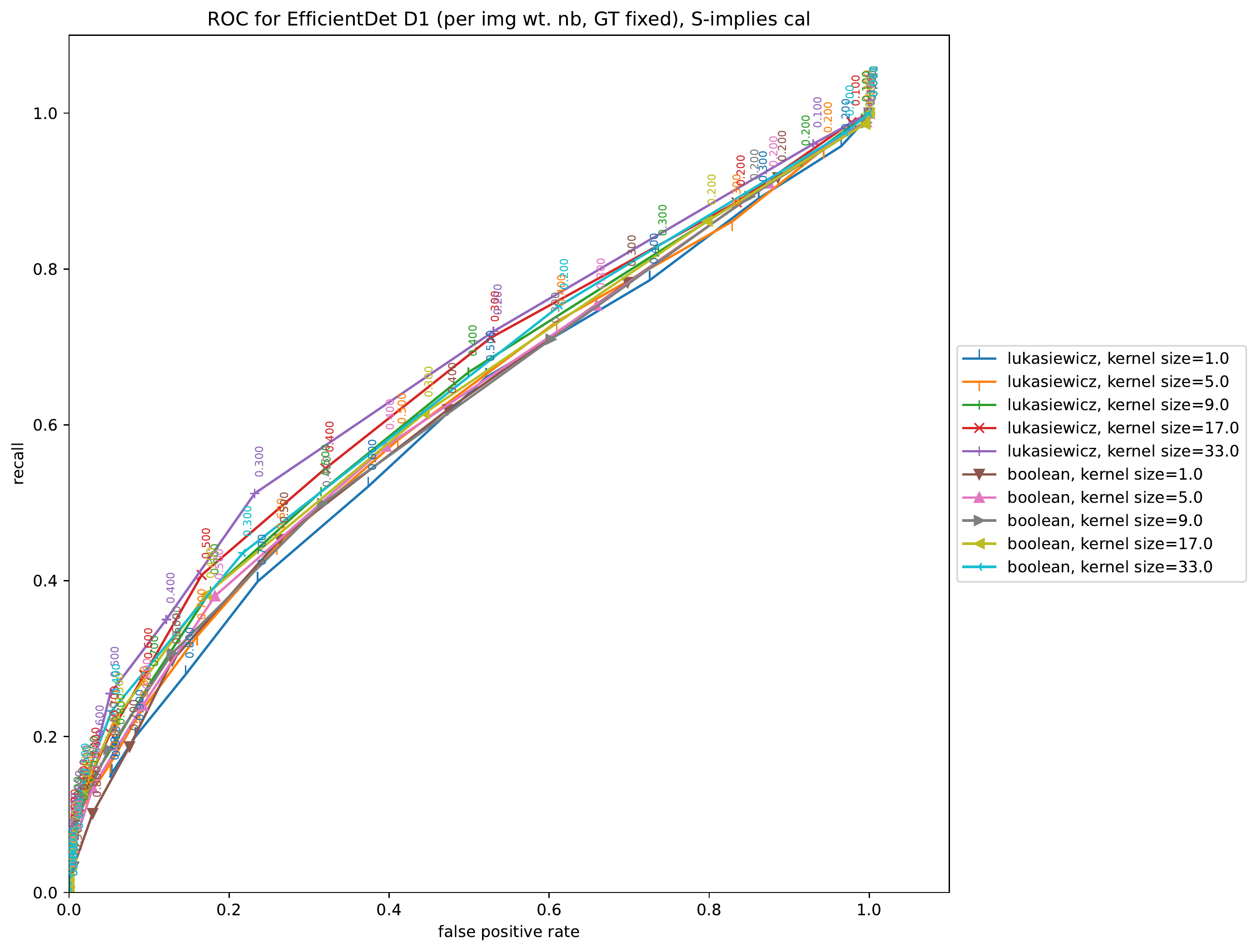}%
    }
    \caption{
    Results for fixed $\constant{ksize}_\text{GT}=9$ and $\thresh{GTreg}^\text{peaks}=0.7$.
    \emph{Top:} Precision-recall curves, \emph{bottom:} ROC curves over $\thresh{reg}^\text{peaks}$,
    \emph{left:} Mask~R-CNN, \emph{right:} EfficientDet~D1.
    }
    \label{fig:perimgnbhoodcurvess-impliesgtfixed}
\end{figure*}

\begin{figure*}
    \centering
    \mbox{
    \includegraphics[width=.6\textwidth]{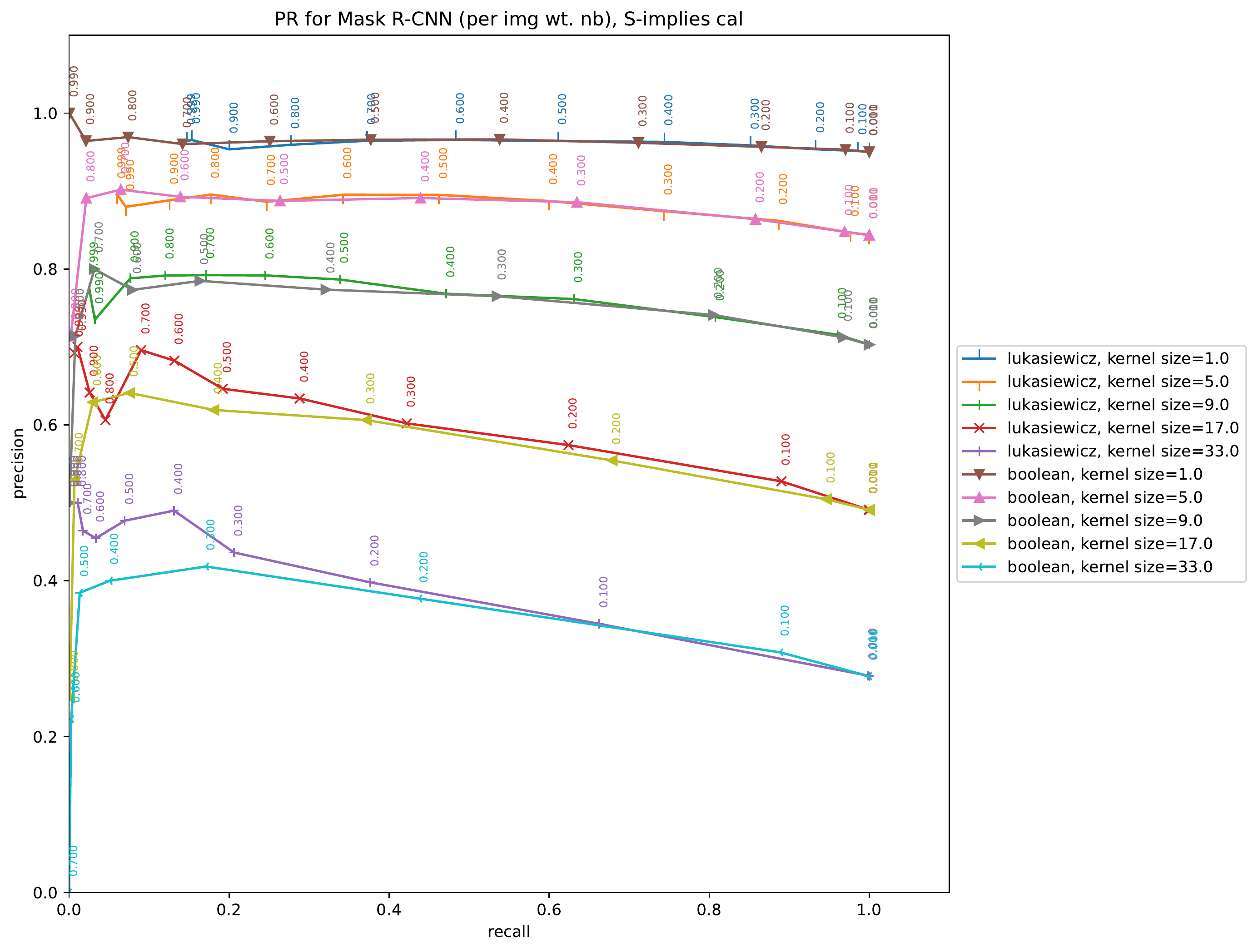}%
    \hspace*{-0.18\textwidth}%
    \includegraphics[width=.6\textwidth]{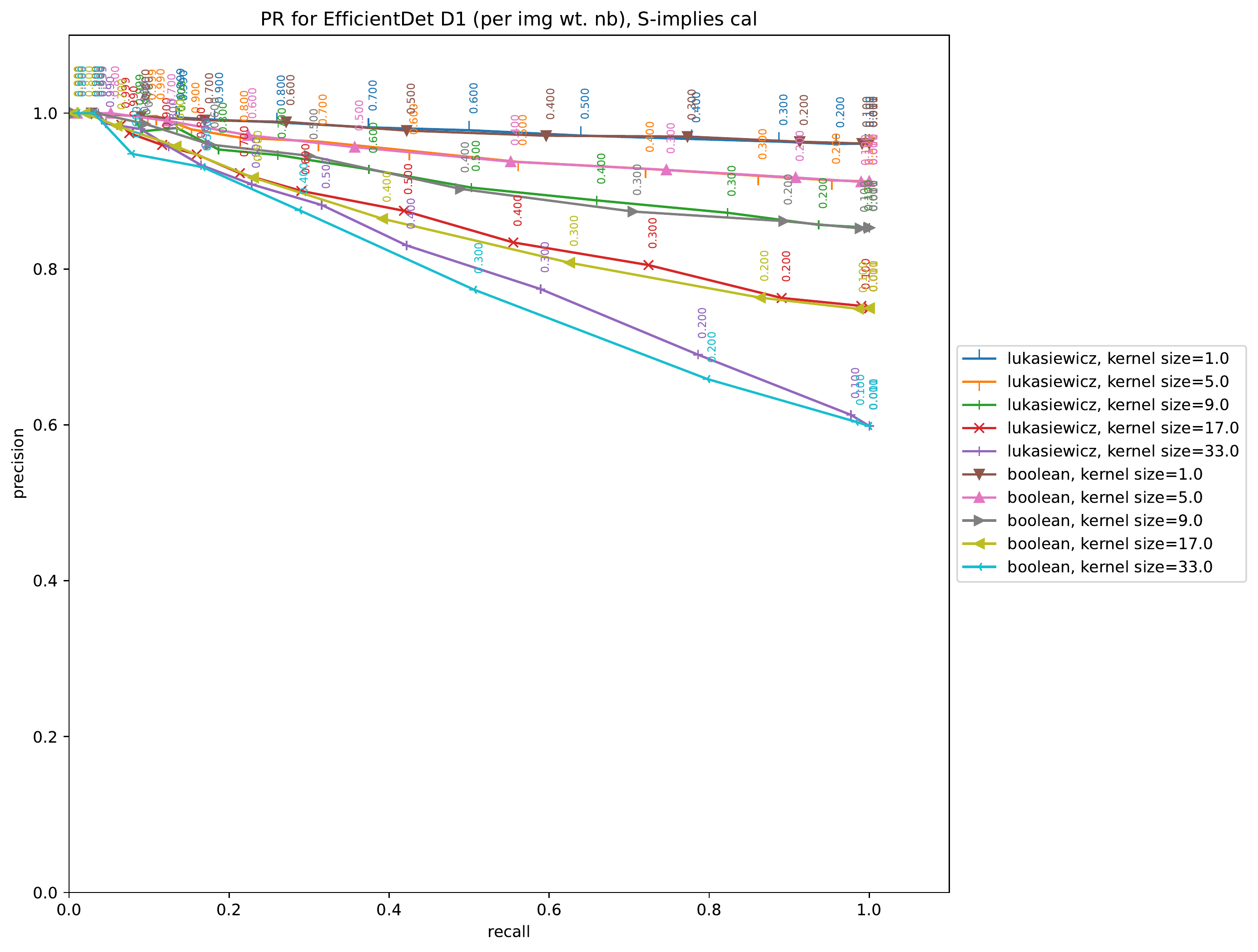}%
    }\\%
    \mbox{
    \includegraphics[width=.6\textwidth]{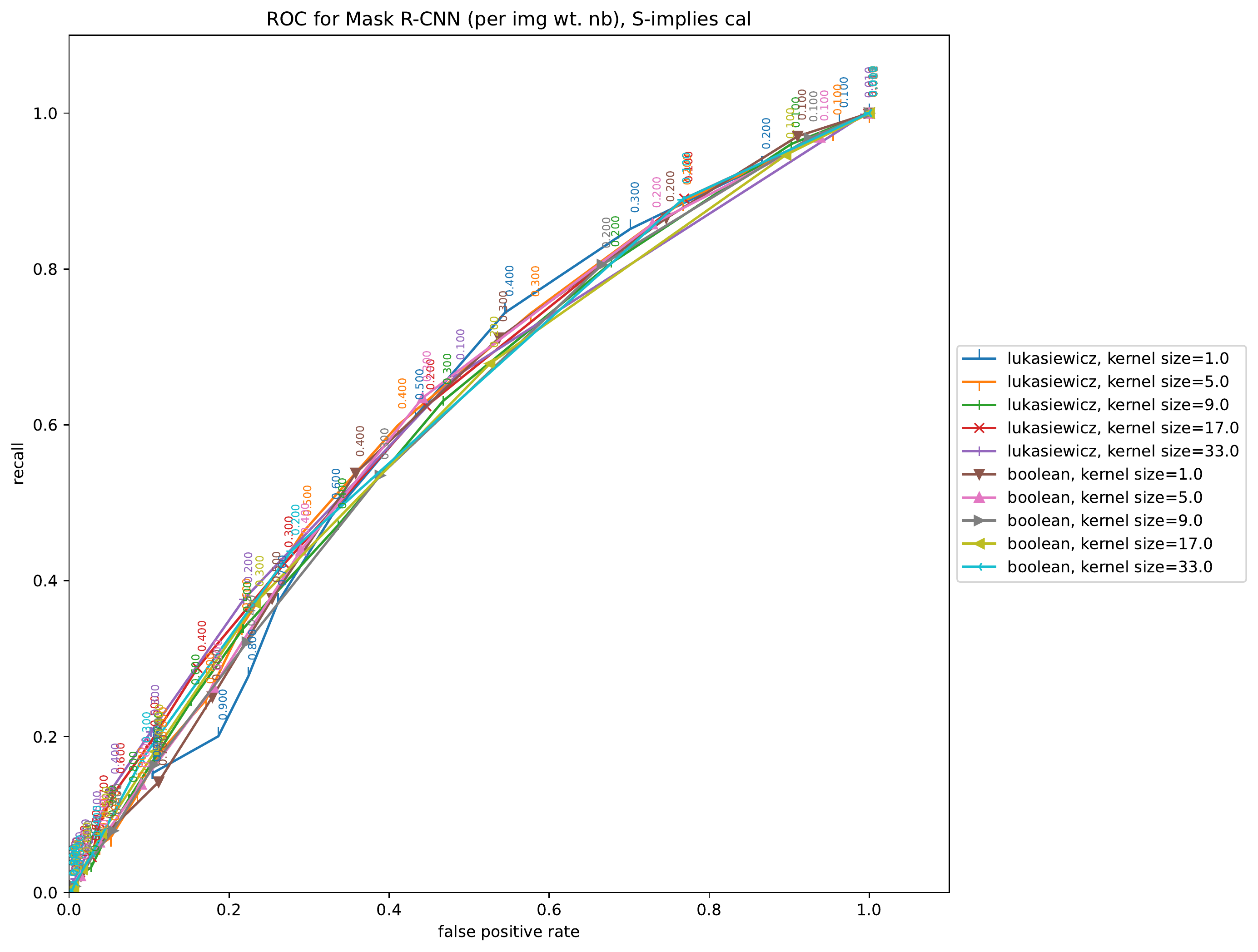}%
    \hspace*{-0.18\textwidth}%
    \includegraphics[width=.6\textwidth]{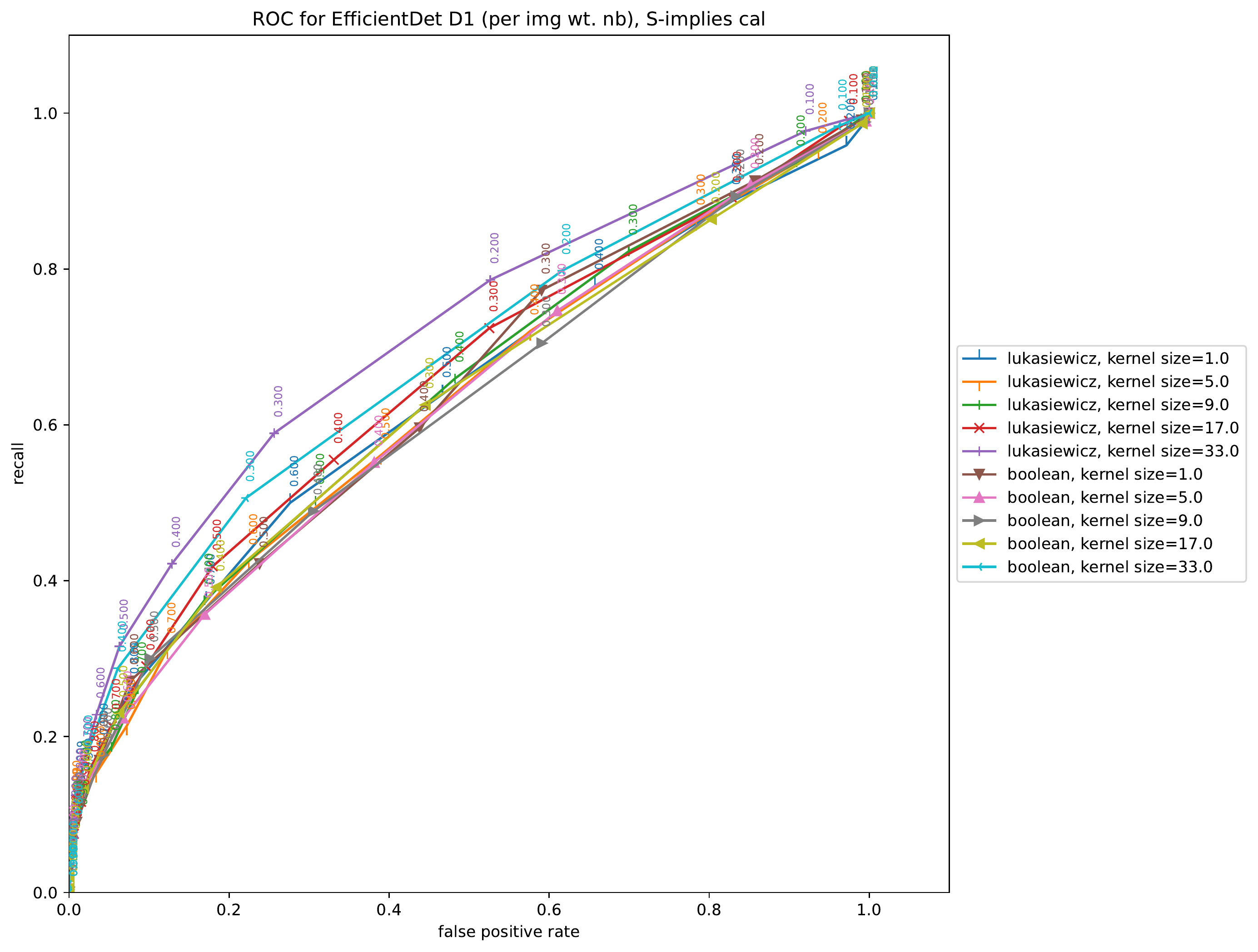}%
    }
    \caption{
    Results for $\constant{ksize}_M=\constant{ksize}_\text{GT}$ and $\thresh{GTreg}^\text{peaks}=0.5$.
    \emph{Top:} Precision-recall curves, \emph{bottom:} ROC curves over $\thresh{reg}^\text{peaks}$,
    \emph{left:} Mask~R-CNN, \emph{right:} EfficientDet~D1.
    }
    \label{fig:perimgnbhoodcurvess-impliesgtadaptive}
\end{figure*}

\begin{figure*}
    \centering
    \mbox{
    \includegraphics[width=.6\textwidth]{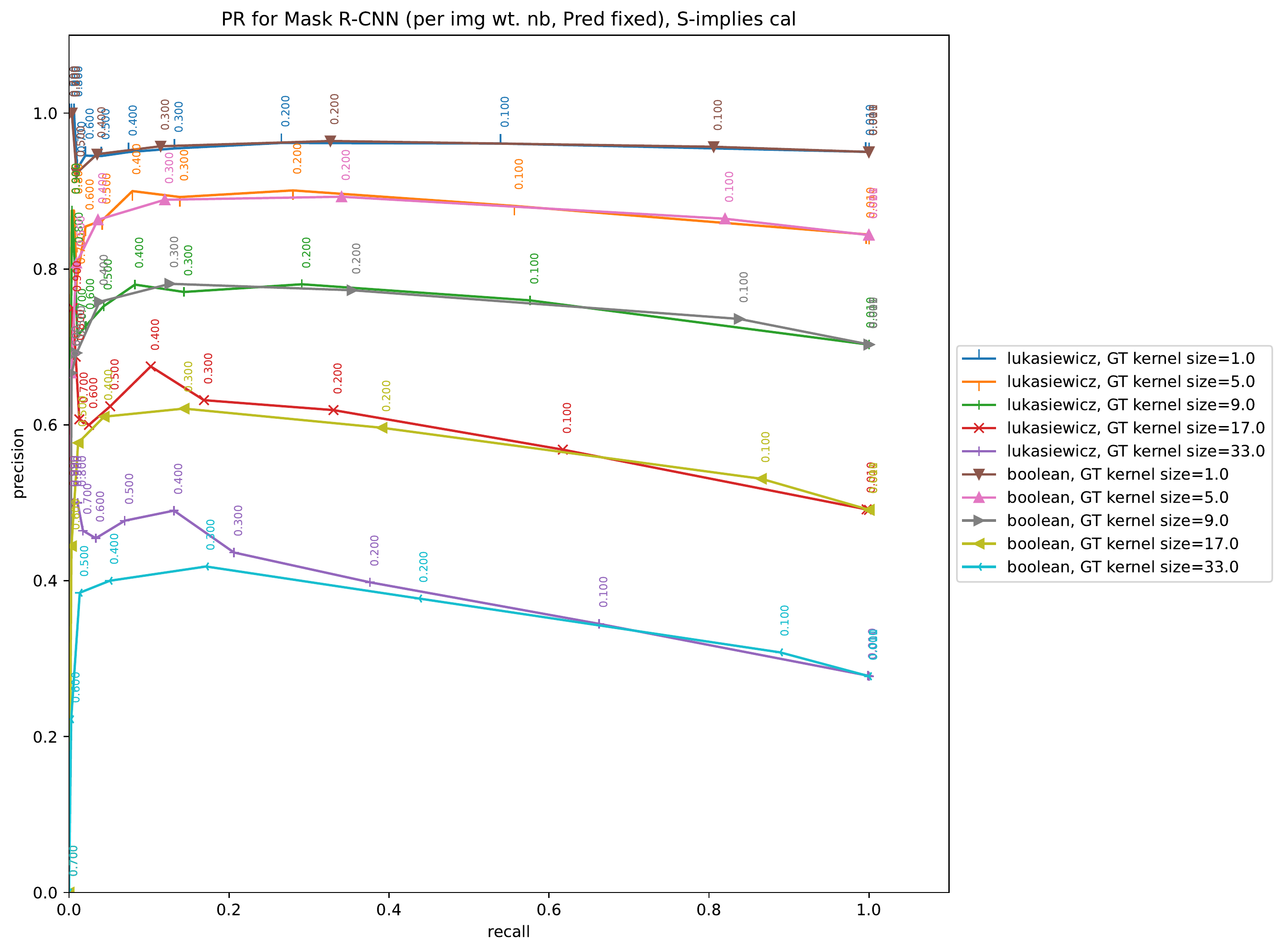}%
    \hspace*{-0.18\textwidth}%
    \includegraphics[width=.6\textwidth]{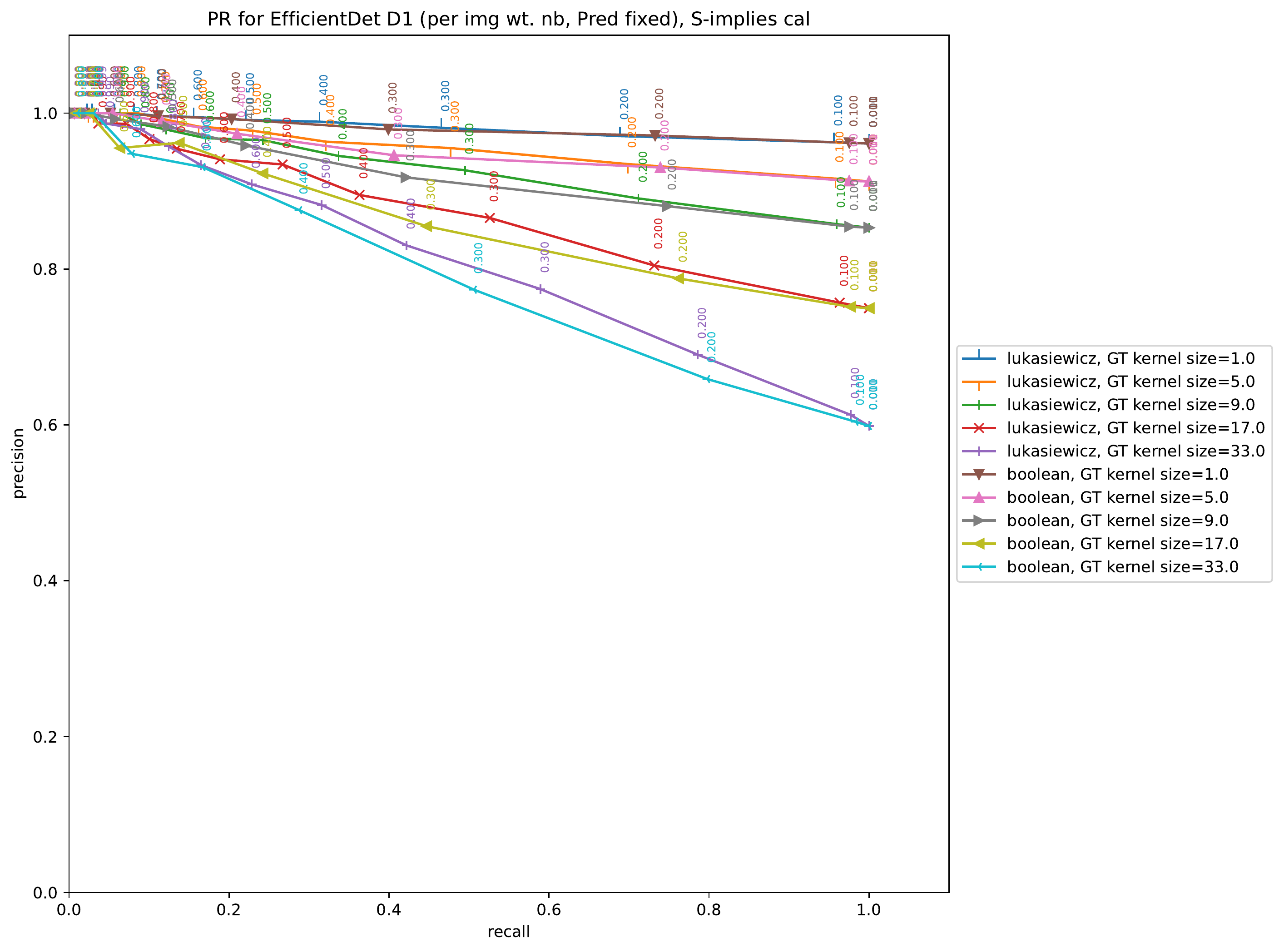}%
    }\\%
    \mbox{
    \includegraphics[width=.6\textwidth]{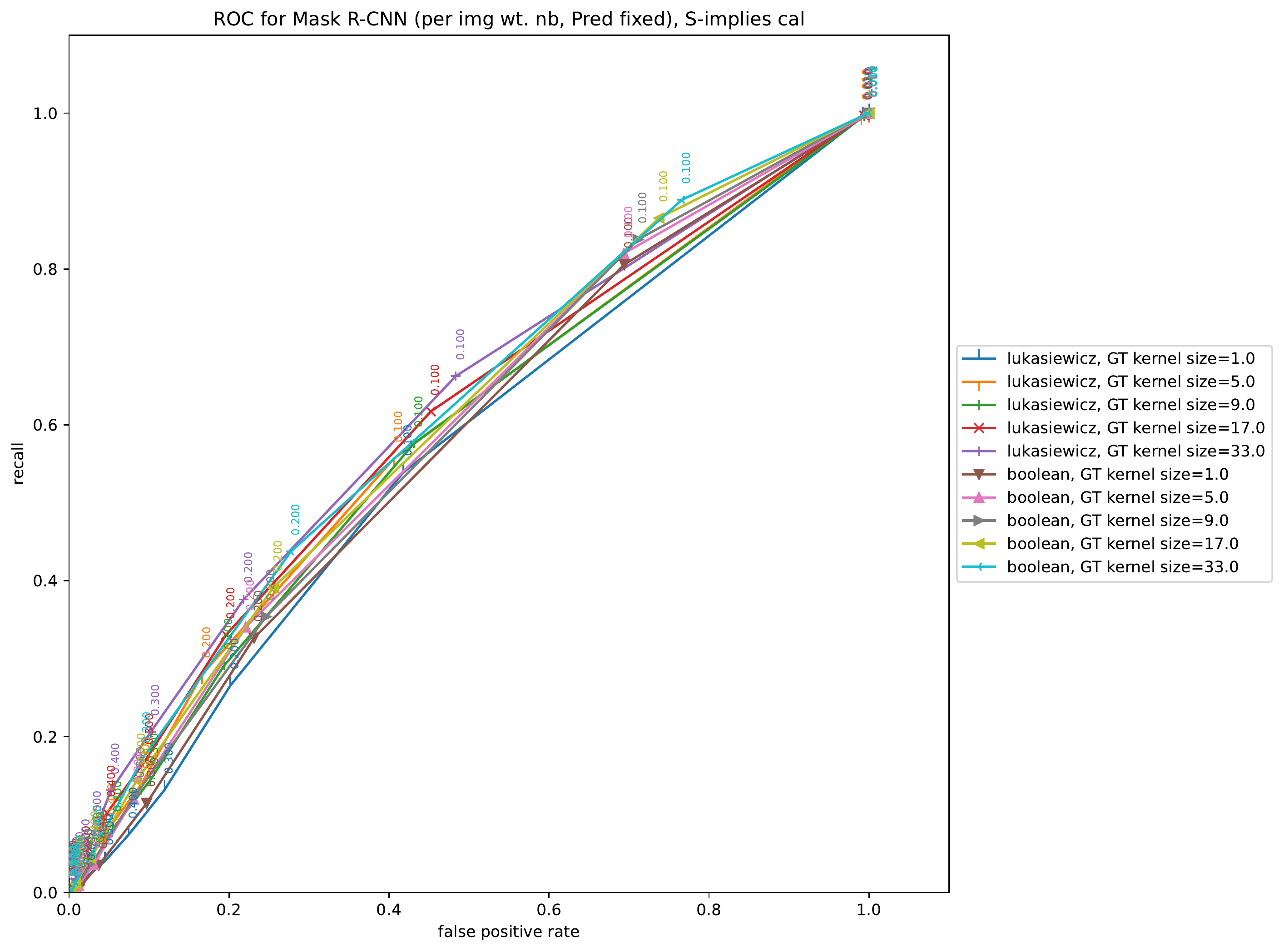}%
    \hspace*{-0.18\textwidth}%
    \includegraphics[width=.6\textwidth]{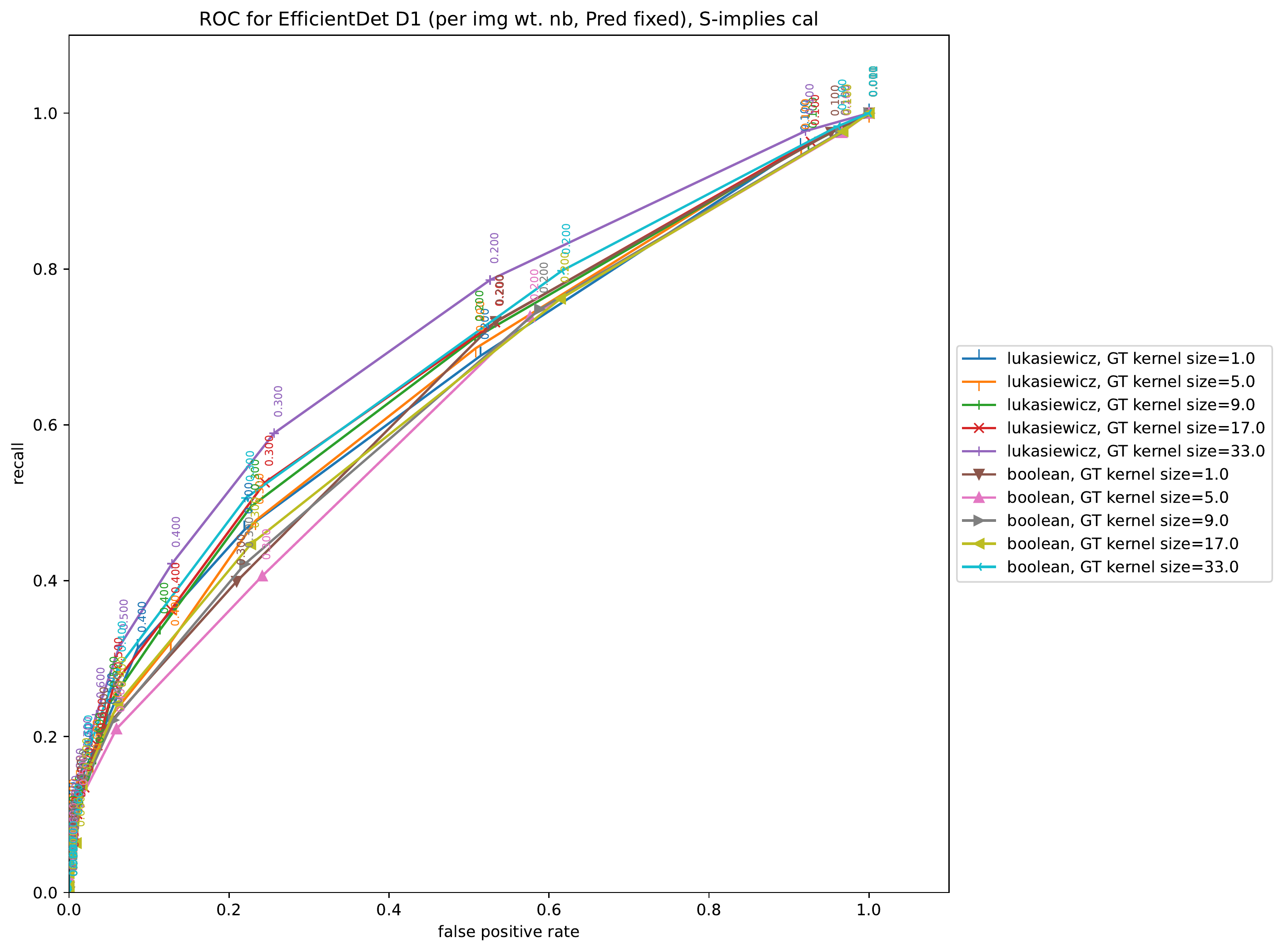}%
    }
    \caption{
    Results for fixed $\constant{ksize}_M=33$ and $\thresh{GTreg}^\text{peaks}=0.5$.
    \emph{Top:} Precision-recall curves, \emph{bottom:} ROC curves over $\thresh{reg}^\text{peaks}$,
    \emph{left:} Mask~R-CNN, \emph{right:} EfficientDet~D1.
    }
    \label{fig:perimgnbhoodcurvess-impliespredfixed}
\end{figure*}

\paragraph{Dependence of Monitor Results on $\constant{ksize}_\text{GT}$}
The following results show that larger kernel sizes of the monitor (prediction) bring some benefit,
regardless of the ground truth kernel size.
Investigated were odd kernel sizes $2^{i}+1, 2\leq i\leq5$ and kernel size of 1 as baseline.
Here evaluated were the neighborhood (nb) image-level formulation of the monitor from \themainpaper{},
for the rule formulation with S-implication and calibration.
Compared were different \L{}ukasiewicz and non-fuzzy logic,
different kernel sizes $\constant{ksize}_M, \constant{ksize}_\text{GT}$,
and binarization thresholds $\thresh{reg}^\text{peaks}, \thresh{GTreg}^\text{peaks}$.
Results are shown
\begin{itemize}[nosep]
    \item for fixed middle sized $\constant{ksize}_\text{GT}=9$ in
        \cref{fig:perimgnbhoodcurvess-impliesgtfixed},
    \item for $\constant{ksize}_M$ fixed in
        \cref{fig:perimgnbhoodcurvess-impliespredfixed},
    \item for $\constant{ksize}_M = \constant{ksize}_\text{GT}$ in
        \cref{fig:perimgnbhoodcurvess-impliesgtadaptive}.
\end{itemize}
The sampling rate of the plotted curves increases towards the value boundaries of $\thresh{reg}^\text{peaks}$.
The results show:
\begin{enumerate}[label=(\arabic*), font=\itshape]
    \item \textbf{Varying $\constant{ksize}_\text{GT}$ for a fixed $\constant{ksize}_M$,
        and the other way round, both have little influence on the performance.}
    \item \textbf{Larger $\constant{ksize}_M$ have slight performance benefits} for common fixed $\constant{ksize}_\text{GT}$.
        This may indicate a better noise robustness of larger kernel sizes.
        Larger kernel sizes can differentiate the quality of truth peaks in a larger range of peak sizes, boosting the influence of large interesting areas while decreasing that of small ones.
        Therefore, in \themainpaper{} we used the largest considered kernel size of 33, which also nicely fits typical body part proportions in the test dataset.
    \item Fuzziness shows a consistent performance benefit (cf.~precision-recall curves).
    \item Increasing $\constant{ksize}_\text{GT}$ decreases the monitor precision.
        This is expected, as a higher $\constant{ksize}_\text{GT}$ also decreases the positive rate (cf.~\cref{tab:perimggtpositiverate}).
\end{enumerate}

\begin{table}
    \centering
    \caption{
    Percentage of the 2693 MS~COCO val2017 images with positive image-level ground truth
    annotations $\bPred{GT}_{\text{reg}}^\text{peaks}$ for different choices of $\constant{ksize}_\text{GT}$
    and $\thresh{GTreg}^\text{peaks}$.
    }
    \label{tab:perimggtpositiverate}
    \begin{tabular}{@{}S S S@{}}
        \toprule
        {$\constant{ksize}_\text{GT}$}
        & {$\thresh{GTreg}^\text{peaks}$}
        & {GT Pos. [\si{\percent}]} \\
        \midrule[\heavyrulewidth]
        1   & 0.5 & 0.95024 \\
        5   & 0.5 & 0.84367 \\
        9   & 0.5 & 0.70293 \\
        9   & 0.7 & 0.59859 \\
        17  & 0.5 & 0.49090 \\
        33  & 0.5 & 0.27739 \\
        \bottomrule
    \end{tabular}
\end{table}

\end{document}